Yau's Computer Award 2019

# Differentially Private M-band Wavelet-Based Mechanisms in Machine Learning Environments

Author: Kenneth Choi
Assistant: Tony Lee

Danbury Math Academy

Danbury, Connecticut, United States of America

Advisor: Xiaodi Wang

August 2019

# Differentially Private M-band Wavelet Based Mechanisms in Machine Learning Environments

Kenneth Choi

Tony Lee

**Abstract**

In the post-industrial world, data science and analytics have gained paramount importance regarding digital data privacy. Improper methods of establishing privacy for accessible datasets can compromise large amounts of user data even if the adversary has a small amount of preliminary knowledge of a user. Many researchers have been developing high-level privacy-preserving mechanisms that also retain the statistical integrity of the data to apply to machine learning. Recent developments of differential privacy, such as those in [14], [16], [20], [30], [34], and [4], drastically decrease the probability that an adversary can distinguish the elements in a data set and thus extract user information. In this paper, we develop three privacy-preserving mechanisms with the discrete M-band wavelet transform that embed noise into data. The first two methods (*LS* and *LS+*) add noise through a "Laplace-Sigmoid" distribution that multiplies Laplace-distributed values with the sigmoid function, and the third method utilizes pseudo-quantum steganography to embed noise into the data. We then show that our mechanisms successfully retain both differential privacy and learnability through statistical analysis in various machine learning environments.

**Highlights**

In this paper, we create three different input perturbation stochastic mechanisms that add or embed noise to sensitive datasets. Our mechanisms improve upon traditional noise addition methods, such as the Laplace mechanism and exponential mechanism mentioned in [14], by using the discrete M-band wavelet transform (DMWT) to convert the dataset into a wavelet domain before adding noise. For the first two mechanisms, we combine the Laplace distribution and the sigmoid function to create a complex stochastic function, and we optimize the mechanisms based on the size of the dataset. In the third mechanism, we propose the use of pseudo-quantum steganography to embed noise into a dataset. Due to the nature of the pseudo-quantum signal, the noisy dataset has an extremely low probability of being correctly denoised by an adversary. While our proposed mechanisms preserve $\varepsilon$-differential privacy, they also maintain the statistical integrity of the datasets. Using five different supervised machine learning environments—logistic regression, support vector machine, support vector regression, classical artificial neural networks, and deep learning—the mechanisms achieve high accuracies in binary classification across multiple datasets. Moreover, our (pseudo-) quantum mechanism is one of the first to use higher computational power to add noise to private data. As data privacy becomes an extremely important issue in our world, and as quantum computing emerges as a major field, our research can link the two branches and shine a light on what data privacy could potentially look like in the future.

**Table of Contents**

# 1  Introduction

More in today's world than ever, there is extreme tension between the mass collection of people's data by corporations and the people's rights to privacy for their personal data. Especially on social media, user data are constantly collected, and a single mishandling of the data pool can lead to the jeopardization of millions of people's data. Recent examples are the Facebook $5 billion FTC fine [17] and the Equifax data breach settlement [15].

There is an important tradeoff between data privacy and statistical analysis. Companies and their research units often rely on user-submitted data for making their products better suited for the market. Users not only wish to preserve anonymity when submitting data but also to be reassured that their individual data cannot be easily identified from a data pool.

If a person has submitted sensitive information in a study whose data has been leaked, an adversary with prior knowledge of the person can use the data to learn something new about the person, and the person's privacy is effectively compromised. For example, a smoker who participates in a survey that requires her to state whether she smokes or not can be harmed if the data is not private. If the study concludes that smoking leads to higher rates of cancer, her insurance rates may be raised if the insurance company finds out she is a smoker from the survey data.

Unfortunately, anonymizing a dataset is not enough to guarantee that someone's information is safely hidden from adversaries. One famous example of a so-called "re-identification attack" is the discovery of Massachusetts governor William Held's medical records in 1997 by a correlation between multiple released datasets [36]: the health database and voter registry. If even anonymization does not protect an individual in a dataset, then what method does?

## 1.1  Prior Works

Differential privacy is a term coined by Dwork *et al.* in [3], and further explained in [12], which sets a privacy budget that quantifies data loss for a mechanism during data release. Over the last decade and a half, researchers have been developing different mechanisms that achieve differential privacy by minimizing the effect of each individual in the dataset. They do this by adding randomly distributed noise to the dataset or a query that slightly obscures the results, enough to ensure privacy but not too much to significantly alter the statistical outcomes.

There are three main ways to add noise to ensure differential privacy, which are input perturbation, which is adding noise to the dataset before a query; objective perturbation, which is adding noise to the objective function in the machine learning model; and output perturbation, which is adding noise to the results after the machine learning model is used.

One well-known differentially private mechanism is the Laplace mechanism [3], which adds Laplace noise with a continuous distribution to a statistical query (output perturbation). However, although it preserves differential privacy, the Laplace mechanism requires the addition of a large amount of noise to both small and large datasets, which hinders the statistical use of the noisy data [33]. Additionally, [33] shows that not adding enough Laplace noise can make the dataset unprotected from a tracker attack.

Another basic mechanism is the exponential mechanism [30], which provides a utilitarian insight into differential privacy by utilizing a quality function to improve upon the usefulness of the Laplace mechanism. The exponential mechanism is also able to apply to non-numeric queries. However, [7] shows that the noise added through the exponential mechanism is asymptotically non-negligible.

Similar to the exponential mechanism, posterior sampling [11] provides a Bayesian approach to differential privacy, and it uses problem-dependent distributions.

Other mechanisms stated in [38], [23], [19], [9], [39], and [6] use composition theorems and the post-processing invariance property stated in [13] to derive mechanisms based on the basic Laplace and exponential mechanisms. Some mechanisms also use alternative definitions of privacy that may relax or tighten its stipulations, such as in [32].

## 1.2 New Ideas

In this research, we use input perturbation to add noise to the dataset directly. We propose three newly derived mechanisms that preserve differential privacy—the first two use a doubly stochastic process that adds Laplace-Sigmoid distributed noise to data, and the third uses pseudo-quantum signals to embed the noise into the data. For all three of our mechanisms, we use discrete 3-band wavelets to transform the data set into the wavelet domain before embedding noise, making the processes impossible to completely reverse without access to the randomized noise.

When first reading our work, one may ask, "What are the benefits of using the discrete M-band wavelet transform?" For one, adding noise to only the approximation part of the transformed matrix, in the case of the first Laplace-Sigmoid mechanism (*LS*), allows the resulting matrix to retain its detailed parts, which are essential for statistical analysis. As a result, we have a transformed matrix that is differentially private yet preserves statistical integrity. Moreover, wavelets effectively scale non-stationary data, which has changing variance and a short-term mean. But, why specifically the M-Band wavelet transform?

Xiao *et al.* in [4] introduces a method of preserving $\varepsilon$-differential privacy via *Privelet* and *Privelet$^+$*, which transform the original dataset via the Haar Wavelet Transform (HWT) and add Laplace noise based on calculated weights. *Privelet* performs well with range-count queries, which solves the inability to produce meaningful results from large aggregate queries using Dwork *et al.*'s mechanism.

However, Privelet's use of the Haar wavelet, the simplest wavelet, can result in easier de-noising of the data by an adversary. Additionally, the Haar wavelet's design is static, unlike M-band wavelets. Most M-band wavelets with $M > 2$ have real values and finite support and are orthogonal, which is a property that is hard to obtain with simpler types of wavelets. M-Band wavelets are also smoother and perform well with image analysis.

Using the discrete M-band wavelet transform, we create two mechanisms that use a combination of Laplace noise and the sigmoid function, and one mechanism that utilizes pseudo-quantum signals to input noise into data (pseudo-quantum steganography).

For the first two mechanisms, the combination of the Laplace distribution and the sigmoid function allows the added noise to be dependent on the dataset, not just a single distribution, similar to the Report Noisy Max mechanism in [20]. The addition of the sigmoid function also adds another degree of randomization that does not add to the privacy budget, since the function is bounded.

For the third mechanism, we propose a new method of treating the noise as a watermark, or secret information, to the data instead of a simple additive distribution. Steganography is the technique of masking information behind seemingly innocent text or images. Following principles of steganography, we develop a mechanism that transforms data into pseudo-quantum signals, which are highly difficult to detect by adversaries, such as in [26]. Although we do not have access to true quantum computers, we are able to use pseudo-quantum methods that mimic qubits. As a result, the mechanism is able to encrypt the noise into an indistinguishable signal, then embed the signal in the discrete wavelet domain. Because the mechanism includes stochastic elements and does not store any values, the key to decrypt the noise is impossible to be retrieved from an adversary.

One such paper that uses a similar model, especially with the use of quantum computing, is [34]. However, [34] only adds Laplace noise to queries, which has the same disadvantages as the Laplace mechanism proposed in [3]: the noise is too large in magnitude for small datasets. The magnitude of the noise in our pseudo-quantum mechanism is able to be controlled by a noise embedding factor. In our tests, we also compare the accuracy of Convolutional Neural Networks with the MNIST image dataset after our pseudo-quantum mechanism with the results from Abadi *et. al.*'s differentially private SGD algorithm in [5].

## 2 Background
### 2.1 Discrete M-band Wavelet Transform (DMWT)

In numerical and function analysis, wavelet transforms decompose an input signal into different frequency levels. The wavelets are discretely sampled and capture both frequency and location information.

Discrete M-band wavelet transforms (DMWT) use M filter banks, where M ≥ 2, to break a K-dimensional signal into M frequency levels. In our paper, we use a slightly modified version of DMWT, choosing to use the signal in multiple column vectors after it is transformed once. In other words, we use the product *WX*, where *W* is the M-band wavelet transform matrix and *X* is the input dataset. The resulting frequency levels include the low-pass approximation (scaling) matrix and M - 1 high-pass detail matrices.

One application of wavelets is in Multiresolution Analysis, shown in [28]. The low-pass filter bank forms linearly independent vectors that span the approximation spaces $V_i$, while the high-pass filter banks form the detail spaces $W_i$. As transformation continues, the approximation spaces are decomposed as the direct sum of higher-level approximation and detail subspaces: $V_i = V_{i+1} \oplus W_{i+1}$.

For instance, the Daubechies 4 wavelet approximation space can be decomposed as $\mathbb{R}^{16} = V_0 = V_2 \oplus W_1 \oplus W_2 \oplus W_3$. When $i = 3$, the subspaces have dimension 2; when $i = 2$, the subspace has dimension 4; and when $i = 1$, the subspace has dimension 8. In the case a 3-band wavelet transform, $V_i = V_{i+1} \oplus W_{i+1,1} \oplus W_{i+1,2}$, and the space can be decomposed as $\mathbb{R}^9 = V_0 = V_2 \oplus W_{2,1} \oplus W_{2,2} \oplus W_{1,1} \oplus W_{1,2}$.

Each M-band orthonormal wavelet has M filter banks. Let an M-band wavelet have filter banks $\alpha^{(1)}, \beta^{(1)}, \dots, \beta^{(M-1)}$. Then the filter banks have the properties for $m = 1, \dots, M - 1$:

$$\|\alpha_i\| = \|\beta^{(1)}\| = \ldots = \|\beta^{(M-1)}\| = 1 \quad (1)$$

$$\sum_{i=1}^{N} \alpha_i = \sqrt{M}, \quad \sum_{i=1}^{N} \beta_i^{(m)} = 0 \quad (2)$$

$$\alpha \cdot \beta^{(m)} = 0 \quad (3)$$

where *N* is the length of each filter bank. Additionally, if the M-band wavelet is also k-regular, it has a fourth property:

$$\sum_{i=1}^{N} i^j \cdot \beta_i^{(m)} = 0 \quad (4)$$

for *j* = 0, ... , *K* - 1.

For all three of our mechanisms, we use a 2-regular 3-band (k = 2, M = 3) orthonormal wavelet transform to break down our input data into the frequency levels. We obtain the 3-band filter banks from [24].

| $\alpha$ | $\beta^{(1)}$ | $\beta^{(2)}$ |
|---|---|---|
| $\alpha_1$ = 0.33838609728386 | $\beta_1^{(1)}$ = -0.11737701613483 | $\beta_1^{(2)}$ = 0.40363686892892 |
| $\alpha_2$ = 0.53083618701374 | $\beta_2^{(1)}$ = 0.54433105395181 | $\beta_2^{(2)}$ = -0.62853936105471 |
| $\alpha_3$ = 0.72328627674361 | $\beta_3^{(1)}$ = -0.01870574735313 | $\beta_3^{(2)}$ = 0.46060475252131 |
| $\alpha_4$ = 0.23896417190576 | $\beta_4^{(1)}$ = -0.69911956479289 | $\beta_4^{(2)}$ = -0.40363686892892 |
| $\alpha_5$ = 0.04651408217589 | $\beta_5^{(1)}$ = -0.13608276348796 | $\beta_5^{(2)}$ = -0.07856742013185 |
| $\alpha_6$ = -0.14593600755399 | $\beta_6^{(1)}$ = 0.42695403781698 | $\beta_6^{(2)}$ = 0.24650202866523 |

Table 2.1: Filter banks for our paper

$$\begin{bmatrix}
\alpha_1 & \alpha_2 & \alpha_3 & \alpha_4 & \alpha_5 & \alpha_6 & 0 & \cdots & & & & & & & & & & & & & & & & & & & 0 \\
0 & 0 & 0 & \alpha_1 & \alpha_2 & \alpha_3 & \alpha_4 & \alpha_5 & \alpha_6 & 0 & \cdots & & & & & & & & & & & & & & & & 0 \\
& & & & & & & & & & \vdots & & & & & & & & & & & & & & & & \\
\alpha_4 & \alpha_5 & \alpha_6 & 0 & \cdots & & & & & & & & & & & & & & & & & 0 & \alpha_1 & \alpha_2 & \alpha_3 \\
\beta_1^{(1)} & \beta_2^{(1)} & \beta_3^{(1)} & \beta_4^{(1)} & \beta_5^{(1)} & \beta_6^{(1)} & 0 & \cdots & & & & & & & & & & & & & & & & & & & 0 \\
& & & & & & & & & & \vdots & & & & & & & & & & & & & & & & \\
\beta_4^{(1)} & \beta_5^{(1)} & \beta_6^{(1)} & 0 & \cdots & & & & & & & & & & & & & & & & & 0 & \beta_1^{(1)} & \beta_2^{(1)} & \beta_3^{(1)} \\
\beta_1^{(2)} & \beta_2^{(2)} & \beta_3^{(2)} & \beta_4^{(2)} & \beta_5^{(2)} & \beta_6^{(2)} & 0 & \cdots & & & & & & & & & & & & & & & & & & & 0 \\
& & & & & & & & & & \vdots & & & & & & & & & & & & & & & & \\
\beta_4^{(2)} & \beta_5^{(2)} & \beta_6^{(2)} & 0 & \cdots & & & & & & & & & & & & & & & & & 0 & \beta_1^{(2)} & \beta_2^{(2)} & \beta_3^{(2)}
\end{bmatrix}$$

Figure 2.1: Example of a 2-regular 3-band 27 x 27 wavelet transform matrix

## 2.2 Quantum Computing and Pseudo-Quantum Signals

Normal computers store information in classical bits, which exist in one of two states: 0 or 1. In contrast, quantum computers use quantum bits, or qubits, which can be in any linear combinations of $|0\rangle = [0, 1]^T$ and $|1\rangle = [1, 0]^T$ as states. Qubit states can be expressed as

$$|\psi\rangle = a|0\rangle + b|1\rangle$$

where $a$ and $b$ are complex numbers and $|a|^2 + |b|^2 = 1$. Qubits also represent points on a 3-D unit sphere. As there are infinite points on a unit sphere, there are infinite states for a qubit.

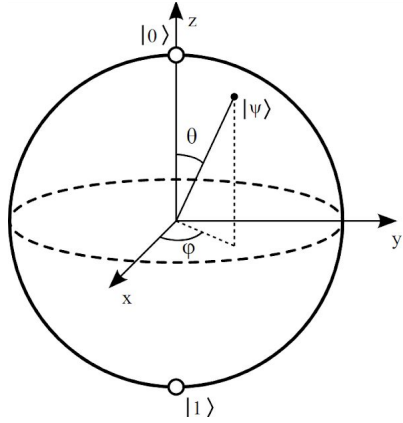
Figure 2.2: The qubit

Another way to write the state of the qubit is

$$|\psi\rangle = \cos\frac{\theta}{2}|0\rangle + e^{i\varphi}\sin\frac{\theta}{2}|1\rangle$$

where $\theta$ and $\varphi$ are real numbers.

As shown in [26], signal $S = [s_1\ s_2\ \ldots\ s_N]^T$ can be transformed into a pseudo-quantum signal with a transformation $F$, where $F$ is called a pseudo quantum signal converter. Let $F$ be a linear transformation where

$$F(s^*) = \frac{m\pi}{3} \text{ and } F(s_*) = \frac{n\pi}{6}$$

$$s^* = \max(s_i),\ s_* = \min(s_i),\ m \in \mathbb{N} \text{ and } n \in \mathbb{N}$$

Then, $F$ transforms signal $S$ into the interval $\left[\frac{m\pi}{6}, \frac{n\pi}{3}\right]$ with $\theta_i = F(s_i)$ for $i = 1, 2, \ldots, N$. (In our paper, we use the case where $m = 1$ and $n = 1$.) The transformed signal values are changed into angles $\theta_i$, and $\theta_i$ can be used to redefine a pseudo qubit as

$$|s_i\rangle = \cos(\theta_i)|0\rangle + \sin(\theta_i)|1\rangle$$

In the case of a pseudo qubit, $\phi = 0$ and the signal can take on values from a circle instead of the sphere shown in Figure 2.2. The transformation result, while not a real quantum signal, is a pseudo-quantum angle that can be computed by classical computers. Thus, pseudo-quantum signals are important for situations when only classical computers can be used to simulate quantum signals. In this research, we propose a new privacy-preserving pseudo-quantum steganography mechanism.

# 3 Privacy-Preserving Mechanisms
## 3.1 ε-differential privacy

In 2006, Dwork, McSherry, Nissim, and Smith's article [3] introduced the concept of *ε*-differential privacy, a mathematical definition for the privacy loss associated with any data release drawn from a statistical database.

**Definition 1** *Let D and D' be neighboring datasets, i.e. that they differ in only one element. A randomized mechanism M satisfies ε-differential privacy if, for any outputs t,*

$$\frac{\Pr[M(D) = t]}{\Pr[M(D') = t]} \leq \exp(\epsilon)$$

For small values of ε, exp (ε) ≈ ε + 1.

ε-differential privacy guarantees that there is low probability for an adversary to discover new information about a unique individual in the dataset, despite having known prior outside knowledge. This fact means the dataset is robust to post-processing. In other words, the adversary is no more likely to pick the true values than if they were to guess randomly.

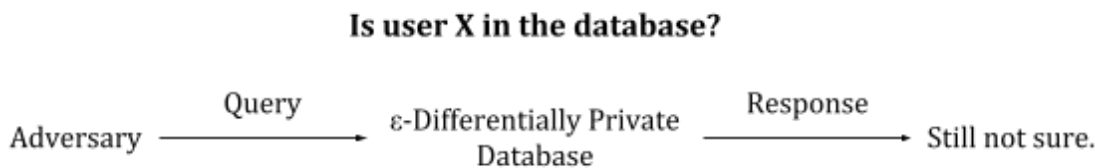

A privacy-preserving mechanism also ensures that the inclusion or removal of one data sample will not alter the overall data set significantly. Individuals who are reluctant to submit data in fear that an adversary will identify their information will be assured that their participation has a marginal effect on the output. It is guaranteed that an ε-differentially private result would not be significantly affected regardless of any one individual's truthful participation. However, in order for the overall data to be useful for statistical analysis, ε needs to be scaled accordingly, i.e. each individual data sample needs to have a non-zero impact.

**Definition 2** *Let $f : D^n \to \mathbb{R}^k$, and let D and D' be neighboring datasets. The sensitivity S(f) of f is defined to be*

$$S(f) = \max_{D, D' \in D^n} \|f(D) - f(D')\|_1$$

*where $\|\cdot\|_1$ is the $\ell_1$ norm.*

For randomized mechanisms *A*, the sensitivity is at most 1, since the neighboring datasets differ in at most 1 element. Dwork *et al.* showed in [8] that for any function $f(x) = \Sigma_i x_i$ in a query $q = f(x) + Y$, where $Y$ is a random variable from the Laplace distribution with mean 0 and scale $1/\varepsilon$, this mechanism is $\varepsilon$-differentially private. However, Dwork *et al.*'s output perturbation Laplace mechanism adds too much noise to even small datasets.

## 3.2 LS Mechanism
### 3.2.1 Steps to Embed LS Noise
Step 1: Discrete M-band Wavelet Transform
Perform the discrete M-band wavelet transform on $D$ to obtain the respective approximation and detail parts of the signal. Although the approximation and detail parts are actually composed of individual column vectors representing each sample, we combine the vectors into the corresponding approximation matrix $A$ and detail matrices $d_i^{(1)}$ for $i = 1, ..., k$.

$$WD = \begin{bmatrix} A \\ d_1^{(1)} \\ \vdots \\ d_k^{(1)} \end{bmatrix}$$

where $k = M - 1$. In the case of 3-Band wavelets, $k = 2$.

Step 2: Data-Sensitive Bound Creation
Define $\mu$ to be the maximum value of all elements in $A$, and $v$ to be the minimum value of all elements in $A$. Let $\gamma$ be the data-sensitive bound parameter. Then, define a new matrix $A^*$ by transforming each element of $A$ to a value bound in the interval $[-\gamma, \gamma]$ through a linear function $g$. Hence, $A^*$ has the property of being sensitive to the values of the original dataset $D$, and the noise added later is, therefore, fitting for $D$. For $\gamma > 0$, we create $A^*$ by the following procedure:

$$\mu = \max(A_{ij}), \quad v = \min(A_{ij})$$

$$A_{ij}^* = g(A_{ij}) = \frac{\gamma(2A_{ij} - \mu - v)}{\mu - v}$$

Step 3: Laplace-Sigmoid Distribution
**Definition 3** *A variable $N_{ij}(y)$ from the Laplace-Sigmoid distribution is defined to be*

$$N_{ij}(y) = \begin{cases} (1 - S(y)) X_{ij} & \text{if } X_{ij} \geq 0 \\ (S(y)) X_{ij} & \text{if } X_{ij} < 0 \end{cases}$$

where $X \sim Lap(0, 1/\varepsilon')$ for some $\varepsilon' > 0$, and $S(y) = \frac{1}{1+e^{-y}}$ is the sigmoid function.

Like *X*, the noise *N* has mean 0.

We can create a final noise matrix $N(A^*)$ by using the Laplace-Sigmoid distribution with $\varepsilon' = \varepsilon/(1 + e^{-\gamma})$. This hybrid noise-generating mechanism uses the shrunken data values obtained in Step 2 and the values from the Laplace-Sigmoid distribution. We can achieve a roughly equal distribution of positive and negative noise values, as the mean of *N* is 0. The new noise matrix *N* is sensitive to the values of the original dataset.

Step 4: Insert New Approximation Matrix

We construct a new approximation matrix $\widehat{A} = A + N(A^*)$ and insert it back into the wavelet transformed data. Because the wavelet matrix *W* is orthogonal, it is easy to transform the transformed data back to its original domain since $W^T = W^{-1}$. We obtain the "noisy" dataset.

$$\widehat{D} = W^T \begin{bmatrix} \widehat{A} \\ d_1^{(1)} \\ \vdots \\ d_k^{(1)} \end{bmatrix}$$

$\widehat{D}$ is now differentially private (we prove it below with Theorem 1) yet it can still be used with relatively little error in machine learning environments. Experiments in various machine learning environments are covered in Section 4. Note also that because we use binary classification in several machine learning environments, we must have discrete values for the last column (labels) in the transformed data. If the label $y_i \geq 0.5$, then we set $y_i = 1$, and if $y_i < 0.5$, then we set $y_i = 0$. The following Lemma 1 and Theorem 1 prove that *LS* is $\varepsilon$-differentially private.

**Lemma 1**  *Let $f(D) = D$ and $F(D) = f(D) + Y$, where $Y \sim Lap\ (0,\ 1/\varepsilon)$. Let $A(X)$ be a randomized mechanism. Then, $A(F(D))$ satisfies $\varepsilon$-differential privacy.*

*Proof:*
Dwork *et al.* prove in [3] that for any randomized function $F$ such that $F(D) = f(D) + Y$, where $Y \sim \text{Lap}\,(0,\ S(F)/\varepsilon)$, $F(D)$ is $\varepsilon$-differentially private.

For emphasis, we restate their proof here.
Let a randomized mechanism $F: D^n \to \mathbb{R}^k$ and an arbitrary point $t \in \mathbb{R}^k$. Then, we can take the ratio of the probability density functions of the Laplace noise added to neighboring datasets *D* and *D'*:

$$\frac{\Pr[F(D) = t]}{\Pr[F(D') = t]} = \prod_{i=1}^{k} \left( \frac{\exp(-\frac{\epsilon |F(D)_i - t_i|}{S(F)})}{\exp(-\frac{\epsilon |F(D')_i - t_i|}{S(F)})} \right)$$

$$= \prod_{i=1}^{k} \left( \exp\left( \frac{\epsilon}{S(F)} |F(D)_i - t_i| - |F(D')_i - t_i| \right) \right)$$

$$\leq \prod_{i=1}^{k} \left( \exp\left( \frac{\epsilon}{S(F)} |F(D)_i - F(D')_i| \right) \right)$$

$$= \exp\left( \frac{\epsilon}{S(F)} \|F(D)_i - F(D')_i\|_1 \right)$$

$$= \exp\left( \frac{\epsilon}{S(F)} S(F) \right)$$

$$= \exp(\epsilon)$$

Thus, by the rules of composability stated in [14], $A(F(D))$ satisfies $\varepsilon$-differential privacy. ∎

**Theorem 1** *Let LS(D) be the LS mechanism with original dataset D and Laplace noise $Y \sim$ Lap $(0, 1/\varepsilon')$, let W be the M-band wavelet transform matrix, and let $\gamma$ be the data-sensitive bound parameter. Then, LS(D) preserves $\varepsilon$-differential privacy, where $\varepsilon' = \varepsilon / (1 + e^{-\gamma})$.*

*Proof:*
We can write the *LS* mechanism as

$$LS(D) = W^T(WD + N') = f(D) + W^T N',$$

$$\text{where } N' = \begin{bmatrix} N \\ 0 \\ \vdots \\ 0 \end{bmatrix}, \text{ and } f(D) = D$$

Since $W^T$ is an orthonormal M-band wavelet, we see that

$$\|W^T N'\| = \sqrt{\langle W^T N', W^T N'\rangle}$$
$$= \sqrt{(W^T N')^T W^T N'}$$
$$= \sqrt{(N')^T W W^T N'}$$
$$= \sqrt{(N')^T N'}$$
$$= \sqrt{\|N'\|^2} = \|N'\|$$

Thus, multiplying by $W^T$ preserves the norm of $N'$:

$$\|W^T N'\| = \|N'\| = \|N\|$$

We also know that $-\gamma \le A_{ij}^* \le \gamma$ by the data-sensitive bound creation linear transformation. Then, we conclude that

$$\max\left(1 - S(A_{ij}^*)\right) = \max\left(S(A_{ij}^*)\right) = \frac{1}{1+e^{-\gamma}},$$
$$\min\left(1 - S(A_{ij}^*)\right) = \min\left(S(A_{ij}^*)\right) = \frac{1}{1+e^{\gamma}}$$

Thus,

$$\left|\frac{N_{ij}}{X_{ij}}\right| = 1 - S\left(A_{ij}^*\right) \text{ or } S(A_{ij}^*) \le \frac{1}{1+e^{-\gamma}}$$

It follows that $(1+e^{-\gamma})|N_{ij}| \le |X_{ij}|$. By Lemma 1, a randomized mechanism $A(F(D)) = f(D) + Y$, where $Y \sim \text{Lap}(0, 1/\varepsilon')$ satisfies $\varepsilon'$-differential privacy. Therefore, since $\varepsilon = (1+e^{-\gamma})\varepsilon'$, $LS(D)$ is $\varepsilon$-differentially private. ∎

### 3.2.2 De-Noising the LS Dataset
As our mechanism is a function, one should expect that an original dataset would be obtainable by simply applying the inverse transformation on the noisy dataset, which we call $\widehat{D}$:

$$D = \widehat{D} - W^T \begin{bmatrix} N \\ 0 \\ \vdots \\ 0 \end{bmatrix}$$

However, since the noise matrix is randomized based on a Laplace-Sigmoid distribution, it is impossible to regain the exact original data without having access to $N$, which is not stored in the first place. Nevertheless, it is possible to regain an approximation of the original dataset, though not very accurate, by having stored the logical values of Laplace-distributed $X$ from the mechanism. In order words, we can define matrix the same size as the noise matrix that stores 1 if the noise in that position is positive and -1 if the noise in that position is negative. We call this logical matrix trace($X$). An adversary should not have direct access to trace($X$), but instead can create a matrix of random logical values of the same size.

We can construct an approximation of the noise matrix by multiplying trace($X$) by a factor, then use it in the inverse transformation function to obtain $D$. However, the factor does not directly correspond to $X$, as the random noise used to transform the data is not accessible. So, one way to approach de-noising is to test a new factor $r$ that brings us close to the true values of the noise in the original mechanism:

$$D = \hat{D} - W^T \begin{bmatrix} rN \\ 0 \\ \vdots \\ 0 \end{bmatrix}$$

How close of an estimate to the original dataset we can obtain is dependent on the values of $\gamma$ and $\varepsilon$, though the former does not affect the noise as much as the latter. Of course, an adversary would not know both of the values used in the mechanism.

To test this de-noising method, we use the IPUMS$_2$ Dataset mentioned in Section 4.1. To justify our point that obtaining the original dataset is nearly impossible without knowing the mechanism's variables, we take the average of the absolute difference between the original dataset and the dataset obtained by varying $r$. We do this for values of $r$ ranging from -1 to 1 with an increment of 0.1, then we take the minimum of the trials. Finally, we do the entire process for 100 trials, corresponding to the value of $\varepsilon$. We keep $\gamma = 1$ constant. Let $H$ = the average minimum absolute difference between the original $D$ and the de-noised $D$.

| $\varepsilon$ | 0.5 | 1 | 2 | 4 | 8 |
|---|---|---|---|---|---|
| $H$ | 0.4208 | 0.2121 | 0.1044 | 0.0522 | 0.0262 |

As is shown, $H$ approximately directly varies with $1/\varepsilon$, i.e. $H \propto 1/\varepsilon$. Therefore, unless the value of $\varepsilon$ is large and trace($X$) is somehow released to the public, an adversary would not be able to retrieve a close form of the original dataset by using a de-noising method.

## 3.3 The LS+ Mechanism
### 3.3.1 Versus the LS Mechanism

At times, it is not adequate to apply the *LS* mechanism on an entire dataset. One such case is if the dataset is too large, then the wavelet matrix may take too long to generate. We encounter this problem by generating wavelets of dimension $3^{10}$ and upwards, as our computers do not have sufficient memory.

The second such case is if the individual wants to disperse the approximation and detail portions of individual blocks in the database rather than add noise to only one of the whole dataset's approximation or details.

Another case in which the *LS+* mechanism is the better option is when the number of rows of the dataset is not numerically close to the wavelet's size. Additionally, using the *LS+* mechanism allows new entries to be incorporated quicker without having to wait for large amounts of samples. On top of making the dataset dynamic to new samples, the *LS+* mechanism allows for the potential of parallel computing to reduce computational time and complexity.

### 3.3.2 Steps to Embed LS+ Noise

To utilize the *LS+* mechanism with 3-band wavelets, as in this paper, the dataset's number of rows must be divisible by the size of the wavelet matrix. We apply the Laplace-Sigmoid distributed noise and add it to the full matrix.

$$D = \begin{bmatrix} a_{11} & a_{12} & a_{13} & a_{14} & a_{15} & a_{16} & a_{17} & a_{18} & a_{19} \\ a_{21} & a_{22} & a_{23} & a_{24} & a_{25} & a_{26} & a_{27} & a_{28} & a_{29} \\ a_{31} & a_{32} & a_{33} & a_{34} & a_{35} & a_{36} & a_{37} & a_{18} & a_{39} \\ a_{41} & a_{42} & a_{43} & a_{44} & a_{45} & a_{46} & a_{47} & a_{48} & a_{49} \\ a_{51} & a_{52} & a_{53} & a_{54} & a_{55} & a_{56} & a_{57} & a_{58} & a_{59} \\ a_{61} & a_{62} & a_{63} & a_{64} & a_{65} & a_{66} & a_{67} & a_{68} & a_{69} \\ a_{71} & a_{72} & a_{73} & a_{74} & a_{75} & a_{76} & a_{77} & a_{78} & a_{79} \\ a_{81} & a_{82} & a_{83} & a_{84} & a_{85} & a_{86} & a_{87} & a_{88} & a_{89} \\ a_{91} & a_{92} & a_{93} & a_{94} & a_{95} & a_{96} & a_{97} & a_{98} & a_{99} \\ - & - & - & - & - & - & - & - & - \\ b_{11} & b_{12} & b_{13} & b_{14} & b_{15} & b_{16} & b_{17} & b_{18} & b_{19} \\ b_{21} & b_{22} & b_{23} & b_{24} & b_{25} & b_{26} & b_{27} & b_{28} & b_{29} \\ \vdots & \vdots & \vdots & \vdots & \vdots & \vdots & \vdots & \vdots & \vdots \end{bmatrix}$$

$$\parallel$$

$$\begin{bmatrix} D_1 \\ D_2 \\ \vdots \end{bmatrix}$$

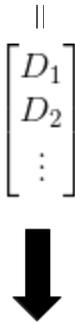

$$\widehat{D}_i = WD_i + N_i \text{ for } i = 1, ..., n$$
where $n$ is the number of blocks

$$N_i \sim \text{Laplace-Sigmoid Distribution}$$

$$f(D_i) = W^T \hat{D}_i = D_i + W^T N_i$$

$$F(D) = \begin{bmatrix} f(D_1) \\ f(D_2) \\ \dots \\ f(D_n) \end{bmatrix} = \begin{bmatrix} D_1 \\ D_2 \\ \dots \\ D_n \end{bmatrix} + \begin{bmatrix} W^T N_1 \\ W^T N_2 \\ \dots \\ W^T N_n \end{bmatrix}$$

We can write the mechanism as a function $F(D) = D + W_n^T N$, where $W_n = \begin{bmatrix} W & \dots & 0 \\ \vdots & \ddots & 0 \\ 0 & 0 & W \end{bmatrix}$ and $N = [N_1\ N_2\ \dots\ N_n]^T$.

The *LS+* mechanism can be proved to be differentially private by Lemma 1 and Theorem 1. It is mathematically similar to the *LS* mechanism, but it is a more complete method of adding noise. Thus, we can use this mechanism for bigger datasets.

Similar to the *LS* mechanism, the transformed dataset $\widehat{D}$ ensures that the $\varepsilon$-differentially private dataset cannot be reversed without much deviation from the original dataset, even with access to the signs of the noise, trace(*X*).

**Corollary 1** *Let n be the number of blocks in dataset D. Then, if each block in the dataset $D^N$ resulting from the LS+ mechanism is $\varepsilon$-differentially private, then F(D) is $\varepsilon$-differentially private.*

*Proof:*
We can apply Theorem 1 to each block of the *LS+* dataset. Let *W* be the wavelet corresponding to each block and letting *N'* be the noise matrix $[N_1\ 0\ \dots\ 0]^T$, we again see that $\|W_n^T N'\| = \|N'\| = \|N_1\| < \|X\|$. For each block, the LS noise added is less than *X*. So, each block is $\varepsilon$-differentially private.

By the Parallel Composition theory in [29], a composition mechanism that involves disjoint subsets of the dataset in each mechanism does not add up in privacy loss for each mechanism; instead, the dataset after the composition mechanism has the maximum $\varepsilon$ value of the submechanisms. Since *LS+* is a composition mechanism whose submechanisms only involve disjoint subsets by partitioning the dataset in blocks, and each block is $\varepsilon$-differentially private, the overall F(D) is $\varepsilon$-differentially private. ∎

## 3.4 The Pseudo-Quantum Mechanism
### 3.4.1 Quantum Steganography
When using a quantum computer, one can generate a true random qubit $S_{ij}$ that satisfies

$$|S_{ij}\rangle = P_{ij_1}|0\rangle + P_{ij_2}|1\rangle$$

to be used in the quantum embedding of the noise into the approximation coefficients. So, each index can be embedded with qubits instead of a pseudo-quantum simulation:

$$\theta_{ij}^E = \begin{cases} \cos^{-1}\left(\cos(\theta_{ij}) + \delta\cos(x_{ij})\right) & \text{if } |P_{ij_1}| \geq |P_{ij_2}| \\ \sin^{-1}\left(\sin(\theta_{ij}) + \delta\sin(x_{ij})\right) & \text{if } |P_{ij_1}| < |P_{ij_2}| \end{cases}$$

However, if one does not have access to a quantum computer, the following pseudo-quantum algorithm must be used.

### 3.4.2 Steps to Embed Pseudo-Quantum Noise
Step 1: Discrete M-band Wavelet Transform
We perform DMWT on the dataset $D$ of size $m$ x $n$ with wavelet matrix $W$. Instead of keeping the approximation part of the dataset in the wavelet domain as a matrix, we split it up into column vectors representing each sample. This mechanism shows the case of a discrete 3-band wavelet transform, which we use in our experiments.

$$WD = \begin{bmatrix} A \\ d_1^{(1)} \\ d_2^{(1)} \end{bmatrix}$$

Step 2: Transform Approximation Coefficients into Angles
In order to embed the noise into the approximation signal, we must transform the approximation coefficients into pseudo-quantum signals, or angles.

$$\theta_{ij} = \frac{\pi(A_{ij} + \mu_1 - 2v_1)}{6(\mu_1 - v_1)} \quad \text{where } A \in M^{i \times j}, \text{ and}$$

$$\text{where } \mu_1 = \max(A_{ij}), \ v_1 = \min(A_{ij})$$

This procedure ensures that the new approximation coefficients $\theta_{ij}$ are bounded in $\left[\frac{\pi}{6}, \frac{\pi}{3}\right]$.

Step 3: Generate Laplace Noise and Transform into Pseudo-Quantum Signals
To make our mechanism stochastic, we randomly generate Laplace-distributed noise with mean 0 and scale $2/\varepsilon$.

$$X_{ij} = \text{Lap}\left(0, \frac{2}{\epsilon}\right)$$

Then, we transform *X* into the same angle bounds as the approximation coefficients. The noise matrix is created to be the same size as the combined approximation matrix, whose number of columns is just *n*. To be added, it must be transformed into pseudo-quantum signals $x_{ij}$:

$$x_{ij} = \frac{\pi(X_{ij} + \mu_2 - 2v_2)}{6(\mu_2 - v_2)} \quad \text{where } X \in M^{i \times j}, \text{ and}$$

$$\text{where } \mu_2 = \max(X_{ij}),\ v_2 = \min(X_{ij})$$

Step 4: Pseudo-Quantum Embedding
We embed the Laplace noise into the transformed approximation coefficients by mimicking a quantum computer's random generation. Using a classical computer, we randomly generate values

$$k_{ij} = \text{rand}\,(0,1)$$

and use the values to embed the noise into a new matrix $\theta^E$ such that

$$\theta^E_{ij} = \begin{cases} \cos^{-1}\left(\cos(\theta_{ij}) + \delta\cos(x_{ij})\right) & \text{if } k_{ij} \geq (1-\eta)/2 \\ \sin^{-1}\left(\sin(\theta_{ij}) + \delta\sin(x_{ij})\right) & \text{if } k_{ij} < (1+\eta)/2 \end{cases}$$

where $\delta$ is the embedding intensity and $\eta$ is the embedding bias, $0 \leq \eta \leq 1$. We use $\delta = 0.1$ and $\eta = 0$ in our trials.

Step 5: Inverse Transformation
In order to obtain the new approximation matrix *A\** with the embedded noise, we do the inverse transformation of the linear transformation before:

$$A^*_{ij} = \frac{6\theta^E_{ij}(\mu_1 - v_1)}{\pi} - \mu_1 + 2v_1$$

Step 6: Inverse DMWT

Finally, we perform the inverse wavelet transform after inserting the new approximation matrix $A^*$.

$$D^* = W^T \begin{bmatrix} A^* \\ d_1^{(1)} \\ d_2^{(1)} \end{bmatrix}$$

Because the dataset is binary, we must have two outputs: 0 and 1. Like in the *LS* and *LS+* mechanisms, if the label $y_i \geq -0.5$, then $y_i = 1$, and if $y_i < -0.5$, then $y_i = 0$.

**Lemma 2**  Let $X_{ij} \sim$ Lap $(0, \sigma)$ and let $x_{ij}$ be the angle in $[\frac{\pi}{6}, \frac{\pi}{3}]$ corresponding to $X_{ij}$. Then, the resulting probability density function of any signal $z$ embedded with $x$ through pseudo-quantum steganography is equal to

$$\frac{\delta}{\sigma\sqrt{2}} e^{-\frac{2|z|}{\sigma}}$$

where $\delta$ is the embedding intensity.

*Proof:*

Let $\theta_{ij}^E$ be the resulting signal of $z$ embedded with $x$. We can write the steganography the same way as in Step 4 of the pseudo-quantum mechanism.

Then, the probability density function $f(y)$ of $\delta \cos(x_{ij})$ can be written as

$$f(y) = \delta \frac{d}{dy} \left( \int_0^{\cos^{-1} y} \frac{1}{\sigma\sqrt{2}} e^{-\frac{2|t|}{2}} dt \right)$$

$$= -\frac{\delta}{\sigma\sqrt{2}(\sqrt{1-y^2})} e^{-\frac{2\cos^{-1}(y)}{\sigma}}$$

where $y \in [\frac{1}{2}, \frac{\sqrt{3}}{2}]$. Hence, the probability density function $p_1(z)$ of $\cos^{-1}(\delta \cos(x_{ij}))$ is given by

$$p_1(z) = \frac{d}{dz} \left( \int_0^{\cos z} -\frac{\delta}{\sigma\sqrt{2}(\sqrt{1-y^2})} e^{-\frac{2\cos^{-1} y}{\sigma}} dy \right)$$

$$= \frac{\delta}{\sigma\sqrt{2}} e^{-\frac{2|z|}{\sigma}}$$

where $z \in [\frac{\pi}{6}, \frac{\pi}{3}]$.

Similarly, assuming $\sin^{-1} y > 0$, the probability density function $g(y)$ of $\delta \sin(x_{ij})$ is

$$g(y) = \delta \frac{d}{dy} \left( \int_0^{\sin^{-1} y} \frac{1}{\sigma \sqrt{2}} e^{-\frac{2|t|}{2}} dt \right)$$

$$= \frac{\delta}{\sigma \sqrt{2}(\sqrt{1-y^2})} e^{-\frac{2\sin^{-1}(y)}{\sigma}}$$

where $y \in [\frac{1}{2}, \frac{\sqrt{3}}{2}]$, and the probability density function $p_2(z)$ of $\sin^{-1}(\delta \sin(x_{ij}))$ is

$$p_2(z) = \frac{d}{dz} \left( \int_0^{\sin z} \frac{\delta}{\sigma \sqrt{2}(\sqrt{1-y^2})} e^{-\frac{2\sin^{-1} y}{\sigma}} dy \right)$$

$$= \frac{\delta}{\sigma \sqrt{2}} e^{-\frac{2|z|}{\sigma}}$$

where $z \in [\frac{\pi}{6}, \frac{\pi}{3}]$. Since $z > 0$, $z = |z|$. Thus, taking the weighted sum of $p_1(z)$ and $p_2(z)$, the probability distribution function $h(z)$ of $\theta_{ij}^E$ can be written as

$$h(z) = \frac{1-\eta}{2} p_1(z) + \frac{1+\eta}{2} p_2(z) = \frac{\delta}{\sigma \sqrt{2}} e^{-\frac{2|z|}{\sigma}} \qquad \blacksquare$$

**Theorem 2** *The pseudo-quantum mechanism with $X_{ij} \sim \text{Lap}(0, 2/\varepsilon)$ is $\varepsilon$-differentially private.*

*Proof:*
The pseudo-quantum mechanism can be written as a function $f$ of $D$:

$$F(D) = W^T \left( WD + \begin{bmatrix} A^* - A \\ 0 \\ \vdots \\ 0 \end{bmatrix} \right)$$

$$= D + W^T \begin{bmatrix} A^* - A \\ 0 \\ \vdots \\ 0 \end{bmatrix}$$

$$\triangleq D + N$$

where *A* is the original approximation matrix and *A\** is the approximation matrix with embedded noise. Using Lemma 2, we can obtain the probability density function of $\theta_{ij}^E$. Then, we see that for neighboring datasets *D* and *D'* and for a query *q*,

$$\begin{aligned}
\frac{\Pr[q(D+N)=R]}{\Pr[q(D'+N')=R]} &= \frac{\Pr[N=q^{-1}(R)-D]}{\Pr[N'=q^{-1}(R)-D']} \\
&= \prod_{i=1}^{m}\prod_{j=1}^{n}\left(\frac{\frac{\delta\epsilon}{2\sqrt{2}}\exp\left(-\epsilon\left|q^{-1}(R)_{ij}-D_{ij}\right|\right)}{\frac{\delta\epsilon}{2\sqrt{2}}\exp\left(-\epsilon\left|q^{-1}(R)_{ij}-D'_{ij}\right|\right)}\right) \\
&= \prod_{i=1}^{m}\prod_{j=1}^{n}\left(\exp\left(\epsilon\left(\left|q^{-1}(R)_{ij}-D'_{ij}\right|-\left|q^{-1}(R)_{ij}-D_{ij}\right|\right)\right)\right) \\
&\leq \exp\left(\epsilon\left\|D-D'\right\|\right) \\
&\leq \exp\left(\epsilon\right)
\end{aligned}$$

Therefore, by Definition 1, the pseudo-quantum mechanism is $\varepsilon$-differentially private. ∎

**Remark 1** *The embedding intensity $\delta$ in the pseudo-quantum mechanism has the least upper bound $\frac{2}{\sqrt{3}}-1$.*

*Proof:*
We know that

$$\theta_{ij}^E = \begin{cases} \cos^{-1}\left(\cos(\theta_{ij})+\delta\cos(x_{ij})\right) & \text{if } k_{ij} \geq (1-\eta)/2 \\ \sin^{-1}\left(\sin(\theta_{ij})+\delta\sin(x_{ij})\right) & \text{if } k_{ij} < (1+\eta)/2 \end{cases}$$

for wavelet approximation pseudo-quantum signals $\theta_{ij}$ and Laplace noise pseudo-quantum signals $x_{ij}$. Since the maximum value of both the cosine and sine functions in the interval $[\frac{\pi}{6},\frac{\pi}{3}]$ is $\sqrt{3}/2$, and since the maximum value of both the inverse cosine and inverse sine functions is 1, we see that

$$\frac{\sqrt{3}}{2}+\delta\frac{\sqrt{3}}{2} \leq 1$$

Therefore, the least upper bound of $\delta$ is $\frac{2}{\sqrt{3}}-1 < 0.155$. This is also shown in [26]. ∎

### 3.4.3 De-Noising the Pseudo-Quantum Dataset

Data encrypted with noise from quantum steganography is impossible to de-noise completely. The noise simply cannot be retrieved without knowing the original dataset, the embedding intensity $\delta$, and the embedding bias $\eta$.

The only way to extract the noise $x$ embedded in the data as pseudo-quantum signals is to use the following formula:

$$x_{ij} = \begin{cases} \cos^{-1}\left(\frac{\cos(\theta_{ij}^E) - \cos(\theta_{ij})}{\delta}\right) & \text{if } k_{ij} \geq (1-\eta)/2 \\ \sin^{-1}\left(\frac{\sin(\theta_{ij}^E) - \sin(\theta_{ij})}{\delta}\right) & \text{if } k_{ij} < (1+\eta)/2 \end{cases}$$

However, extracting the noise is redundant because the adversary would need to have the original approximation coefficients in the first place. Therefore, the transformed dataset is resistant to harmful post-processing.

Moreover, for $\eta = 0$, the probability for an adversary to decode the noise added to the approximation coefficients by randomization in $k_{ij}$ is only $2^{-mn/3}$, where the dataset has size $m \times n$.

## 4 Machine Learning Environments and Experiment Results

In our paper, we employ the three mechanisms in five machine learning environments.

### 4.1 Datasets

We utilize 2 different datasets in our paper depending on the mechanism and the machine learning environment. The first dataset, IPUMS, is obtained from the Integrated Public Use Microdata Series (IPUMS) [1] and has 8 predictors and 1 binary results column with 3190032 instances. The predictors are: number of generations, detailed information for number of generations, family size, race, detailed information for race, whether the participant is deaf, whether the participant has a cognitive disability, and whether the participant is blind. The binary resultant is whether the participant speaks English or not.

Due to the *LS+* mechanism's ability to add noise with small-sized wavelets, we are able to use many samples in the *LS+* training and testing sets. We choose to use 90000 samples for training and 30000 samples for testing, and all samples are randomly chosen from the IPUMS Dataset every time the mechanism is run. For all *LS+* trials, we use blocks of size 9 x 9.

For the *LS* and pseudo-quantum mechanisms, we use 19683 samples for training and 729 samples for testing because our computer only has enough memory to generate and store a maximum of $3^9$ as the length of the 3-band wavelet. The samples for the pseudo-quantum mechanism are randomly chosen from the IPUMS dataset.

The third dataset, MNIST [2], is a public image dataset commonly used in machine learning research. We use it to train and test Convolutional Neural Networks (CNN). MNIST consists of a training set with 60000 images of handwritten digits (0 to 9) and a testing set with 10000 different images of handwritten digits. The sets are pre-randomized. Each image is 28 x 28, but since we must add noise to each image, we cut the last column and last row of each image (which are insignificant to classification) in order to use the pseudo-quantum mechanism with 3-band wavelet length of 27.

|  | LS | LS+ | Pseudo-Quantum |
|---|---|---|---|
| Logistic Regression | IPUMS$_2$ | IPUMS$_1$ | IPUMS$_2$ |
| Support Vector Machine (SVM) | IPUMS$_2$ | IPUMS$_1$ | IPUMS$_2$ |
| Support Vector Regression (SVR) | IPUMS$_2$ | IPUMS$_1$ | IPUMS$_2$ |
| Classical Artificial Neural Network | IPUMS$_2$ | IPUMS$_1$ | IPUMS$_2$ |
| Deep Learning | IPUMS$_2$ | IPUMS$_1$ | MNIST |
| IPUMS$_1$: 90000 samples in training set, 30000 samples in testing set; randomized from 3190032 samples IPUMS$_2$: 2187 samples in training set, 729 samples in testing set; randomized from 3190032 samples MNIST: 60000 samples of 27 x 27 images in training set, 10000 samples of 27 x 27 images in testing set ||||

Note: We train each classifier or network using the noisy training set. Then, we add noise to the testing set predictors and use the classifier or network to predict the testing set responses. To calculate the accuracy, we take the absolute difference between the real responses and the responses predicted by the machine learning model, and we find the proportion of correct predictions. This process makes up one trial. For all the machine learning environments except Deep Learning, we run 1000 trials for each $\varepsilon$ = 0.5, 1, 2, 4, 8, and we take the average accuracy. We only run one trial in Deep Learning for each value of $\varepsilon$ for the three mechanisms due to time restraint and computer resources.

In addition to $\varepsilon$, each mechanism has other variables. For the *LS* and *LS+* mechanisms, we test each $\varepsilon$ value at $\gamma$ = 0.5, 1, 2. For the pseudo-quantum mechanism in machine learning environments, we keep $\delta$ = 0.1 and $\eta$ = 0 constant for all $\varepsilon$ values. For Section 4.6 (Differentially Private Imaging), we vary $\delta$. Note that neither $\delta$ nor $\eta$ affect the privacy of the noisy dataset.

## 4.2  Logistic Regression
Data analysis often requires tools and algorithms to predict future cases based on pre-existing datasets. Machine learning algorithms, such as logistic regression, take pre-existing datasets separated into predictor variables and label variables, analyze the relationship between the two, and create models that classify or predict future samples of data [31].

We use a logistic regression model as the first machine learning environment to test the *LS* mechanism with Database A. Logistic regression uses the logistic curve, which uses the sigmoid function.

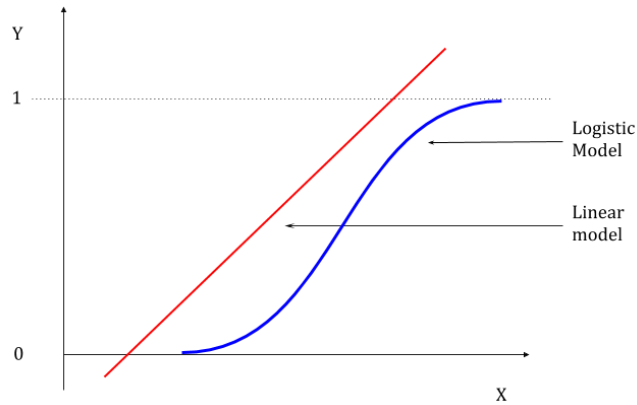

Figure 4.1: Logistic regression model vs. linear regression model

Logistic regression is a classification model represented by $p$, a function of x that represents the probability that the respective label to x is 1. If $p \geq 0.5$, then the model predicts that the x value corresponds to a label value of 1; conversely, if $p < 0.5$, then the model predicts 0 for the label value. The function $p$ for a logistic regression model is defined to have the property:

$$\ln\left(\frac{p}{1-p}\right) = \beta_0 + \beta_1 \cdot x$$

where $x$ is the data vector, $\beta_0$ is the bias constant, and $\beta_1$ is the coefficient vector to the predictor variables. The solution of this equation, $\beta_0 + \beta_1 \cdot x$, is called the decision boundary and divides the two label classes. If $x$ is one dimension, the decision boundary would be a point, and if x is two dimensional, it would be a line [8].

The equation of a logistic regression model is the sigmoid function of the logit function of $p$. The logit function is the natural logarithm of the odds of $p$.

$$\text{odds} = \frac{p}{1-p}$$

$$\text{logit}(p) = \ln(\text{odds}) = \ln\left(\frac{p}{1-p}\right)$$

Substituting the logit function of $p$ for the equation $\beta_0 + \beta_1 \cdot x$, we obtain the form of a logistic regression model to be

$$p(x) = \frac{1}{1 + e^{-(\beta_0 + \beta_1 \cdot x)}}$$

A logistic regression model uses maximum likelihood estimation (MLE) to solve for the parameters $\beta_0$ and $\beta_1$. Let $p$ be a probability function where the probability of the label variable $y = 1$ is $p$, and the probability of $y = 0$ is $1-p$. With predictor variable $X = \{x_1, x_2, ..., x_n\}$ and binary label variable $Y = \{y_1, y_2, ..., y_n\}$, the likelihood function is:

$$L(\beta_0, \beta_1) = \prod_{i=1}^{n} p(x_i)^{y_i} (1 - p(x_i))^{1-y_i}$$

Solving for the maximum of the likelihood function requires finding the derivative, but since the likelihood function is difficult to differentiate, the log-likelihood function is used instead:

$$l(\beta_0, \beta_1) = \sum_{i=1}^{n} y_i \log p(x_i) + (1 - y_i) \log (1 - p(x_i))$$

Simplifying the log-likelihood function gives:

$$l(\beta_0, \beta_1) = \sum_{i=1}^{n} -\log (1 + e^{\beta_0 + x_i \cdot \beta_1}) + \sum_{i=1}^{n} y_i(\beta_0 + x_i \cdot \beta_1)$$

Define the cost function $J(\beta_0, \beta_1)$ as:

$$J(\beta_0, \beta_1) = -l(\beta_0, \beta_1) = \sum_{i=1}^{n} [\log(1 + e^{\beta_0 + x_i \cdot \beta_1}) - y_i(\beta_0 + x_i \cdot \beta_1)]$$

Then to maximize the log-likelihood function is the same as minimizing the cost function. Our aim is to estimate $\beta_0$ and $\beta_1$ so that the cost function is minimized. So we first take partial derivatives of $J(\beta_0, \beta_1)$ with respect to $\beta_0$ and $\beta_{1j}$ to derive the stochastic gradient descent rule:

$$\frac{\partial J}{\partial \beta_0} = \sum_{i=1}^{n} \frac{e^{\beta_0 + x_i \cdot \beta_1}}{1 + e^{\beta_0 + x_i \cdot \beta_1}} - \sum_{i=1}^{n} y_i = \sum_{i=1}^{n} p(x_i) - \sum_{i=1}^{n} y_i$$

$$\text{and } \frac{\partial J}{\partial \beta_{1j}} = \sum_{i=1}^{n} \frac{1}{1 + e^{\beta_0 + x_i \cdot \beta_1}} e^{\beta_0 + x_i \cdot \beta_1} x_{ij} - \sum_{i=1}^{n} y_i x_{ij}$$

$$= \sum_{i=1}^{n} (p(x_i; \beta_0, \beta_1) - y_i) x_{ij}$$

where $j = 1, 2, ..., m$, and $\beta_1 = \begin{bmatrix} \beta_{11} \\ \vdots \\ \beta_{1m} \end{bmatrix}$

Now, in order to find the weights of the model, we take a step proportional to positive direction of the gradient to minimize the cost function. Furthermore, we add coefficients, the learning rates $\eta_1$ and $\eta_0$ to the weight updates.

$$\beta_{1j}(k+1) := \beta_{1j}(k) - \eta_1 \sum_{i=1}^{n}(P(x_i, \beta_0(k), \beta_1(k)) - y_i)x_{ij}, \; j = 1, ..., m$$

$$\beta_0(k+1) := \beta_0(k) - \eta_0 \sum_{i=1}^{n}(P(x_i, \beta_0(k), \beta_1(k)) - y_i)$$

| LS - Logistic Regression | | | | | | | | | | | | | | | |
|---|---|---|---|---|---|---|---|---|---|---|---|---|---|---|---|
| $\gamma$ | 0.5 | 1 | 2 | 0.5 | 1 | 2 | 0.5 | 1 | 2 | 0.5 | 1 | 2 | 0.5 | 1 | 2 |
| $\varepsilon$ | 0.5 | | | 1 | | | 2 | | | 4 | | | 8 | | |
| % Accuracy | 94.99 | 94.81 | 94.87 | 94.88 | 95.08 | 97.65 | 98.68 | 99.41 | 99.40 | 99.96 | 99.93 | 99.89 | 100.0 | 100.0 | 99.99 |

| LS+ - Logistic Regression | | | | | | | | | | | | | | | |
|---|---|---|---|---|---|---|---|---|---|---|---|---|---|---|---|
| $\gamma$ | 0.5 | 1 | 2 | 0.5 | 1 | 2 | 0.5 | 1 | 2 | 0.5 | 1 | 2 | 0.5 | 1 | 2 |
| $\varepsilon$ | 0.5 | | | 1 | | | 2 | | | 4 | | | 8 | | |
| % Accuracy | 94.86 | 94.82 | 94.90 | 94.91 | 95.05 | 97.82 | 98.05 | 99.03 | 99.40 | 99.78 | 99.89 | 99.90 | 99.99 | 99.99 | 99.99 |

| PQ - Logistic | | | | | |
|---|---|---|---|---|---|
| $\varepsilon$ | 0.5 | 1 | 2 | 4 | 8 |
| % Accuracy | 94.80 | 94.92 | 95.12 | 95.19 | 95.39 |

## 4.3 Support Vector Machine and Regression
### 4.3.1 Support Vector Machine (SVM)
The support vector machine is a machine learning algorithm used for classifying data into binary classes. An SVM model takes a training data set consisting of predictors $\{x_1, x_2, ..., x_n\}$ and their respective binary labels $\{y_1, y_2, ..., y_n\}$ where $y \in \{0, 1\}$, and constructs a hyperplane that separates the data based on the labels.

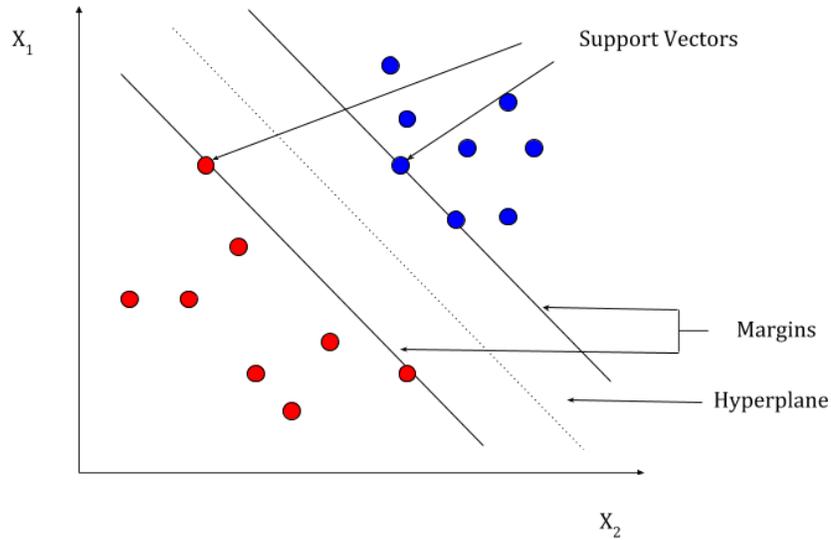

Figure 4.2: Support vector machine model with binary targets

Labels y are classified from predictors *x* and parameters *w* and *b*:

$$y = \text{sign}(w^T x + b)$$

SVM models achieve high classification accuracy by maximizing the width between the margins, or the distance between the hyperplane and the closest vectors on each side (the "support vectors") [18]. With small margins, the hyperplane may overfit the training data on either side and make classification errors with actual testing data. For a linear SVM model, the margin width is maximized by solving the following primal problem:

$$\min \frac{1}{2} \|w\|^2 + C \sum_{i=1}^{n} \xi_i$$
$$\text{subject to: } y_i(w \cdot x_i + b) \geq 1 - \xi_i, \ \xi_i \geq 0$$

where $\xi_i$ is the slack variable that gives the model flexibility for some misclassifications while still maintaining the largest possible margins. $C$ represents the trade-off between misclassification cases and large margins. The solution to this minimization problem is also a stationary point of Lagrange function:

$$L(w, b, \xi, \alpha, \beta) = \frac{1}{2} \|w\|^2 + C \sum_{i=1}^{n} \xi_i - \sum_{i=1}^{n} \alpha_i (y_i(w \cdot x_i + b) - 1 + \xi_i) - \sum_{i=1}^{n} \beta_i \xi_i$$

where $\alpha_i$ and $\beta_i$ are non-zero Lagrange multipliers, as shown in [31]. This Lagrange equation is solved by maximizing the following dual problem:

$$W(\alpha) = \sum_{i=1}^{n} \alpha_i - \frac{1}{2} \sum_{i=1}^{n} \sum_{j=1}^{n} \alpha_i \alpha_j y_i y_j (x_i \cdot x_j)$$

$$\text{subject to:} \sum_{i=1}^{n} y_i \alpha_i = 0, \ 0 \leq \alpha_i \leq C$$

Then, we obtain the weight vector $w$:

$$w = \sum_{i=1}^{n} \alpha_i y_i x_i$$

So the classifier can be written as $f(x) = sgn(\sum_{i=1}^{n} \alpha_i y_i x_i^T x + b)$.

When a linear hyperplane cannot accurately separate the dataset, the SVM model uses a nonlinear function $\varphi(x)$ that maps the inputs in a higher dimension and allows the SVM model to separate data that are impossible to linearly separate in lower-dimensional spaces [18]. Then the corresponding kernel function is defined by $K(x_i, x_j) = \varphi(x_i) \cdot \varphi(x_j)$. The margin for an SVM model using a kernel function $K(x_i, x_j)$ is maximized by solving the primal problem in the transformed space:

$$\min \frac{1}{2} \|w\|^2 + C \sum_{i=1}^{n} \xi_i$$

$$\text{subject to:} \ y_i \left( \sum_{i=1}^{n} \alpha_i y_i K(x_i, x_j) + b \right) \geq 1 - \xi_i, \ \xi_i \geq 0$$

The margin can also be maximized by the corresponding dual problem in the transformed space:

$$W(\alpha) = \sum_{i=1}^{n} \alpha_i - \frac{1}{2} \sum_{i=1}^{n} \sum_{j=1}^{n} \alpha_i \alpha_j y_i y_j K(x_i, x_j)$$

$$\text{subject to:} \sum_{i=1}^{n} y_i \alpha_i = 0, \ 0 \leq \alpha_i \leq C$$

The weight vector for the kernel function optimization problem then becomes:

$$w = \sum_{i=1}^{n} \alpha_i y_i \phi(x_i)$$

And the classifier now is given by $f(x) = sgn(\sum_{i=1}^{n} \alpha_i y_i k(x_i, x) + b)$.

The dual problem in linear SVM and the kernel trick can look complicated, but note that most of the $\alpha_i$ values are equal to 0. For our paper, we use the Gaussian kernel function in the form

$$K(x_i, x_j) = \exp(\frac{-\|x_i - x_j\|^2}{2\sigma^2}), \ \sigma > 0$$

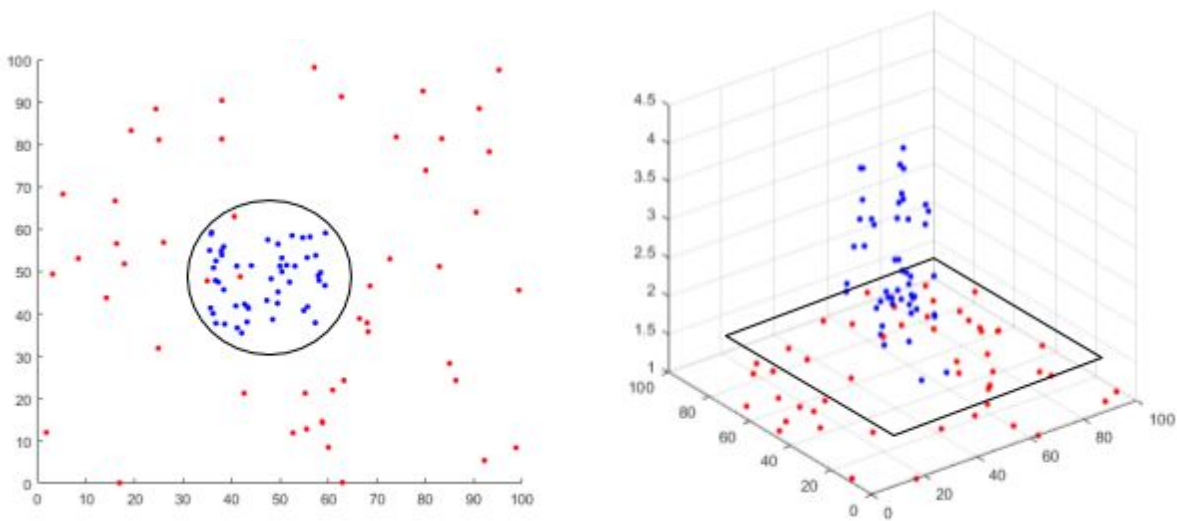

Figure 4.3: Example of data mapped in 3-dimensional space with Gaussian kernel

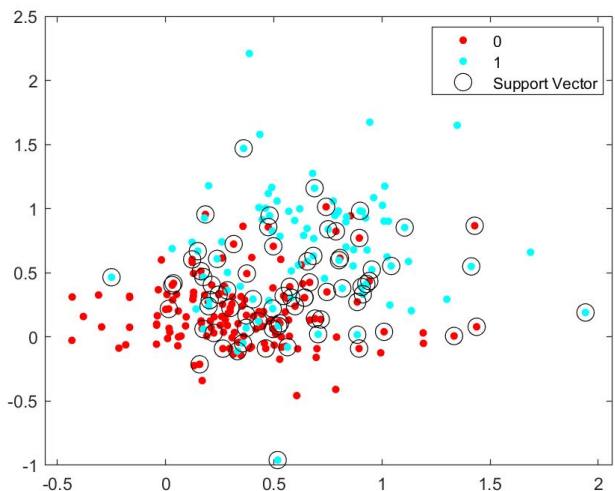

Figure 4.4: Nonlinear support vector machine classification of $\widehat{D}$ with the first two predictors

| LS - SVM | | | | | | | | | | | | | | | |
|---|---|---|---|---|---|---|---|---|---|---|---|---|---|---|---|
| $\gamma$ | 0.5 | 1 | 2 | 0.5 | 1 | 2 | 0.5 | 1 | 2 | 0.5 | 1 | 2 | 0.5 | 1 | 2 |
| $\varepsilon$ | | 0.5 | | | 1 | | | 2 | | | 4 | | | 8 | |
| % Accuracy | 94.90 | 94.88 | 94.90 | 94.86 | 94.90 | 95.06 | 96.60 | 97.06 | 97.22 | 97.87 | 97.96 | 98.03 | 98.26 | 98.17 | 98.29 |

| LS+ - SVM | | | | | | | | | | | | | | | |
|---|---|---|---|---|---|---|---|---|---|---|---|---|---|---|---|
| $\gamma$ | 0.5 | 1 | 2 | 0.5 | 1 | 2 | 0.5 | 1 | 2 | 0.5 | 1 | 2 | 0.5 | 1 | 2 |
| $\varepsilon$ | | 0.5 | | | 1 | | | 2 | | | 4 | | | 8 | |
| % Accuracy | 94.84 | 94.86 | 94.86 | 94.87 | 94.88 | 94.87 | 95.58 | 96.91 | 97.43 | 98.30 | 98.53 | 98.64 | 98.89 | 98.93 | 98.95 |

| PQ - SVM | | | | | |
|---|---|---|---|---|---|
| $\varepsilon$ | 0.5 | 1 | 2 | 4 | 8 |
| % Accuracy | 94.81 | 94.84 | 94.85 | 94.93 | 94.94 |

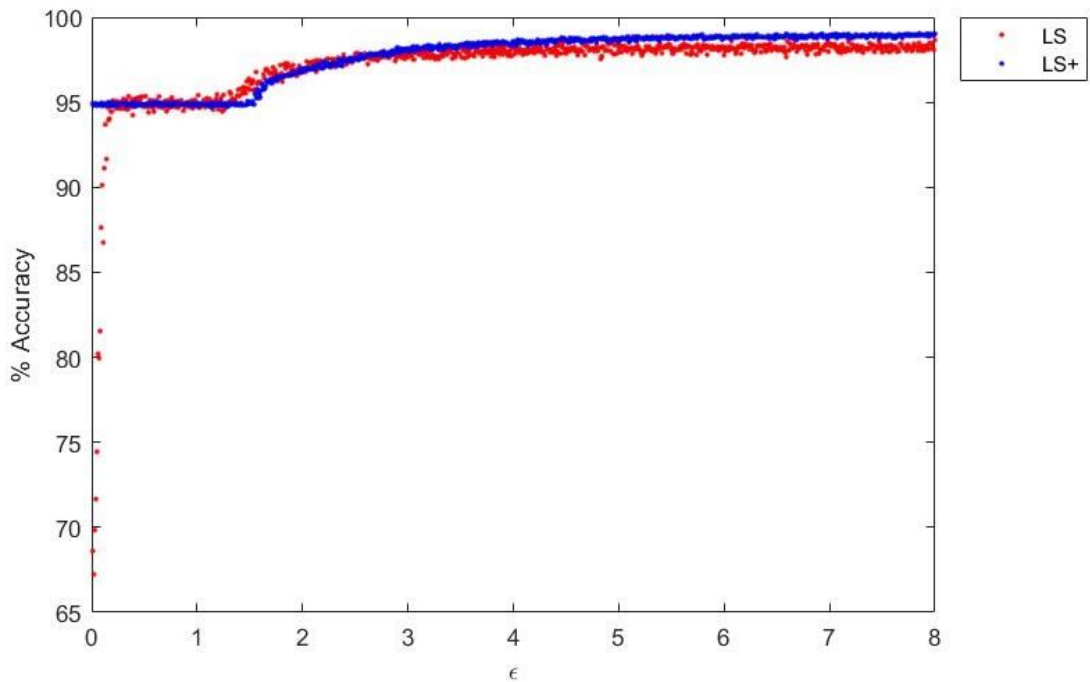

Figure 4.5: Average percent accuracy of SVM classification where $\gamma = 1$

### 4.3.2 Support Vector Regression (SVR)

A support vector model can also be used for regression. Instead of classifying data into binary classes in SVM models, support vector regression uses a training set of predictors $\{x_1, x_2, \ldots, x_n\}$ and their respective label values $\{y_1, y_2, \ldots, y_n\}$, $y \in \mathbb{R}$, to construct a model that attempts to output label values that are within an error bound of $\epsilon$ from the actual observed value (not to be confused with the $\varepsilon$ parameter from $\varepsilon$-differential privacy).

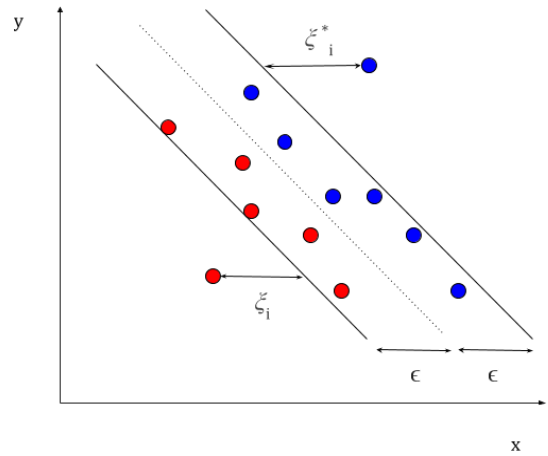

Figure 4.6: Support vector regression model with real targets

SVR differs from SVM by outputting values that are not limited to binary; instead, they range across all real numbers. Therefore, the output $y$ is defined to be:

$$y = \sum_{i=1}^{n}(\alpha_i - \alpha_i^*) \cdot K(x_i, x_j) + b$$

where $K(x_i, x_j)$ is the Gaussian kernel function aforementioned in Section 4.3.1. Since it is impossible for the model to perfectly output points that fall within the $\epsilon$ error bound, slack variables are used to allow errors up to the values of $\xi_i$ and $\xi_i^*$ [35]. The optimization problem for the margin width for an SVR model, which is extremely similar to the optimization equation for SVM due to both using support vector models, then becomes:

$$\min \frac{1}{2}\|w\|^2 + C\sum_{i=1}^{N}(\xi_i + \xi_i^*)$$

$$\text{subject to: } y_i - \left(\sum_{i=1}^{n}(\alpha_i - \alpha_i^*) \cdot K(x_i, x_j) + b\right) \leq \epsilon + \xi_i$$

$$\left(\sum_{i=1}^{n}(\alpha_i - \alpha_i^*) \cdot K(x_i, x_j) + b\right) - y_i \leq \epsilon + \xi_i^*$$

$$\xi_i, \xi_i^* \geq 0$$

In order to find the optimal ϵ value for the *LS* and *LS+* mechanisms, we run 100 trials for each of the ϵ values ranging from 0.000 to 0.500 with increments of 0.001. The accuracies from the 100 trials are averaged for each ϵ value. For larger ϵ values, the error bounds at either side of the SVR model are larger and the model predicts with more errors. Therefore, we pick the smallest ϵ value that corresponds with the highest average accuracy across the 100 trials to create SVR models for the 1000 trials.

For the pseudo-quantum mechanism, we run trials for each ϵ value ranging from 0.250 to 0.500 with increments of 0.005. The ϵ testing range for the pseudo-quantum mechanism is determined by 100 test trials that concludes that the optimal ϵ values are all greater than 0.300 and less than 0.500.

| LS - SVR | | | | | | | | | | | | | | | |
|---|---|---|---|---|---|---|---|---|---|---|---|---|---|---|---|
| $\gamma$ | 0.5 | 1 | 2 | 0.5 | 1 | 2 | 0.5 | 1 | 2 | 0.5 | 1 | 2 | 0.5 | 1 | 2 |
| $\varepsilon$ | 0.5 | | | 1 | | | 2 | | | 4 | | | 8 | | |
| % Accuracy | 94.85 | 94.89 | 94.86 | 94.86 | 94.88 | 94.87 | 94.86 | 94.79 | 94.86 | 94.86 | 94.88 | 94.85 | 94.83 | 94.90 | 94.82 |

| LS+ - SVR | | | | | | | | | | | | | | | |
|---|---|---|---|---|---|---|---|---|---|---|---|---|---|---|---|
| $\gamma$ | 0.5 | 1 | 2 | 0.5 | 1 | 2 | 0.5 | 1 | 2 | 0.5 | 1 | 2 | 0.5 | 1 | 2 |
| $\varepsilon$ | 0.5 | | | 1 | | | 2 | | | 4 | | | 8 | | |
| % Accuracy | 94.87 | 94.87 | 94.87 | 94.87 | 94.87 | 94.89 | 94.84 | 94.87 | 94.86 | 94.88 | 94.87 | 94.87 | 94.87 | 94.87 | 94.89 |

| PQ - SVR | | | | | |
|---|---|---|---|---|---|
| $\varepsilon$ | 0.5 | 1 | 2 | 4 | 8 |
| % Accuracy | 94.8644 | 94.8807 | 94.8815 | 94.8663 | 94.8542 |

## 4.4 Classical Artificial Neural Networks

Neural networks are an excellent tool for modeling and outperform many other traditional machine learning methods. They can also handle large amounts of data similar to Deep Learning, so after feature extraction, we can utilize the classical neural network to the fullest extent.

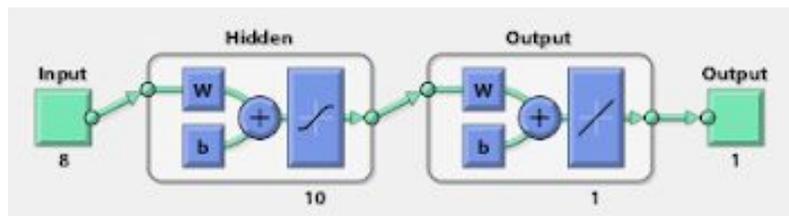

Figure 4.7: We utilize a 2-layer shallow neural network with one hidden layer that contains 10 neurons. Above is an example when softmax outputs 1.

Neural networks consist of an input layer, hidden layers, and an output layer. We use a shallow two-layer feed-forward network with the activation as the sigmoid function:

$$R(z) = \frac{1}{1+e^{-z}}$$

and softmax for the output neurons. To train the network through backpropagation, we use stochastic gradient descent (SGD) provided by the MATLAB Machine Learning toolbox.

### 4.4.1 The Feed-Forward Pass

Our network has parameters $(w, b) = (w^{(1)}, b^{(1)}; \ldots ; w^{(n_l)}, b^{(n_l)})$, where $b_i^{(l)}$ is the bias associated with unit $i$ later $l + 1$, and where $w_{ij}^{(l)}$ denotes the parameter (weight) associated with unit $i$ in layer $l$ and unit $j$ in layer $l + 1$. Then, we can let

$$a_i^{(1)} = x_i, \; i = 1, \ldots, m \text{ and } z_i^{(2)} = \sum_{j=1}^{n} \left(w_{ij}^{(l)} a_j^{(1)}\right) + b_i^{(l)},$$

$$a_i^{(2)} = R\left(z_i^{(2)}\right), \ldots, a_i^{(l)} = R\left(z_i^{(l)}\right)$$

Thus,

$$z_i^{(l+1)} = \sum_{j=1}^{n} \left(w_{ij}^{(l)} a_j^{l}\right) + b_i^{(l)}$$

$$= w_i^{(l)T} a^{(l)} + b_i^{(l)}, \text{ where } a^{(l)} = \begin{bmatrix} a_1^{(l)} \\ \vdots \\ a_n^{(l)} \end{bmatrix}, b_i^{(l)} = \begin{bmatrix} b_{i1}^{(l)} \\ \vdots \\ b_{in}^{(l)} \end{bmatrix} \text{ and } l \geq 1$$

Therefore, for input $z_i^{(l+1)}$, the activation function is

$$R\left(z_i^{(l+1)}\right) = R\left(w_i^{(l)T} a^{(l)} + b_i^{(l)}\right)$$

### 4.4.2 Backpropagation

To train a neural network, we use backpropagation. In our case, we use stochastic gradient descent to find the optimal parameters quickly.

In LMS learning, the distance is given by $\frac{1}{2}\sum_{p}(T_p - O_p)^2$. We must minimize

$$E(w,b) = \frac{1}{n}\sum_{i=1}^{n} \frac{1}{2} \|h_{w,b}(x_i) - y_i\|^2 + \frac{\lambda}{2} \sum_{i,j} \left(w_{ij}^{(l)}\right)^2$$

where $h_{w,b}(x)$ is the output from the neural network.

First, we perform a feed-forward pass, computing the activations at $a^{(2)}$, $a^{(3)}$, ... and up to the output layer $L_{n_l}$, where $n_l$ is the number of layers. For each output at note $i$ in $L_{n_l}$, set

$$\delta_i^{(n_l)} = \frac{\partial E}{\partial \left(z_i^{(n_l)}\right)} = -\left(y_i - a_i^{(n_l)}\right) R'\left(z_i^{(n_l)}\right)$$

Then, for all hidden layers $l = n_l - 1, n_l - 2, ..., 2$, for each node $i$ in layer $l$, set

$$\delta_i^{(l)} = \left(\sum_{j=1}^{l+1} w_{ji}^{(l)} \delta_j^{(l+1)}\right) R'\left(z_i^{(l)}\right), \quad \delta^{(l)} = \left(w^{(l)} \delta^{(l+1)}\right) R'\left(z^{(l)}\right)$$

and compute

$$\frac{\partial E}{\partial w_{ij}^{(l)}} = a_i^{(l)} \delta_i^{(l+1)}, \quad \nabla_{w^{(l)}} E = a^{(l)} \delta^{(l+1)}, \quad \frac{\partial E}{\partial b_i^{(l)}} = \delta_i^{(l+1)}$$

$$\text{and } \nabla_{b^{(l)}} E = \delta^{(l+1)}$$

To implement stochastic gradient descent, set $\Delta_{(0)}^{(l)} = 0$ for all $l$. The increment is 0, so there is no change. Then, for $l = 1$ and $n_1 = 1$, compute

$$\nabla_{w^{(l)}} E(w, b, x_j) \text{ and } \nabla_{b^{(l)}} E(w, b, x_j)$$

To update the gradient, set

$$\Delta w^{(l)}(k+1) = \Delta w^{(l)}(k) + \nabla_{w^{(l)}} E \text{ and } \Delta b^{(l)}(k+1) = \Delta b^{(l)}(k) + \nabla_{b^{(l)}} E$$

Finally, we have

$$w^{(l)}(k+1) = w^{(l)}(k) - \alpha_1 \triangle w^{(l)}(k+1)$$
$$= w^{(l)}(k) - \alpha_1(\triangle w^{(l)}(k) + \nabla_{w^{(l)}} E)$$
$$= w^{(l)}(k) - \alpha_1(\triangle w^{(l)}(k) + a^{(l)}\delta^{(l+1)})$$
$$b^{(l)}(k+1) = b^{(l)}(k) - \alpha_2 \triangle b^{(l)}(k+1)$$
$$= b^{(l)}(k) - \alpha_2(\triangle b^{(l)}(k) + \nabla_{b^{(l)}} E)$$
$$= b^{(l)}(k) - \alpha_2(\triangle b^{(l)}(k) + \delta^{(l+1)})$$

where $\alpha_1$ and $\alpha_2$ are learning rates, respectively.

| | LS - Classical NN | | | | | | | | | | | | | | |
|---|---|---|---|---|---|---|---|---|---|---|---|---|---|---|---|
| $\gamma$ | 0.5 | 1 | 2 | 0.5 | 1 | 2 | 0.5 | 1 | 2 | 0.5 | 1 | 2 | 0.5 | 1 | 2 |
| $\varepsilon$ | 0.5 | | | 1 | | | 2 | | | 4 | | | 8 | | |
| % Accuracy | 94.57 | 94.73 | 95.26 | 95.27 | 96.66 | 98.03 | 99.19 | 99.47 | 99.59 | 99.94 | 99.95 | 99.95 | 100.0 | 100.0 | 100.0 |

| | LS+ - Classical NN | | | | | | | | | | | | | | |
|---|---|---|---|---|---|---|---|---|---|---|---|---|---|---|---|
| $\gamma$ | 0.5 | 1 | 2 | 0.5 | 1 | 2 | 0.5 | 1 | 2 | 0.5 | 1 | 2 | 0.5 | 1 | 2 |
| $\varepsilon$ | 0.5 | | | 1 | | | 2 | | | 4 | | | 8 | | |
| % Accuracy | 94.59 | 94.77 | 95.59 | 95.15 | 97.18 | 98.26 | 98.35 | 99.10 | 99.50 | 99.68 | 99.80 | 99.87 | 99.90 | 99.90 | 99.91 |

| PQ - Classical NN | | | | | |
|---|---|---|---|---|---|
| $\varepsilon$ | 0.5 | 1 | 2 | 4 | 8 |
| % Accuracy | 94.81 | 94.86 | 94.87 | 94.89 | 94.94 |

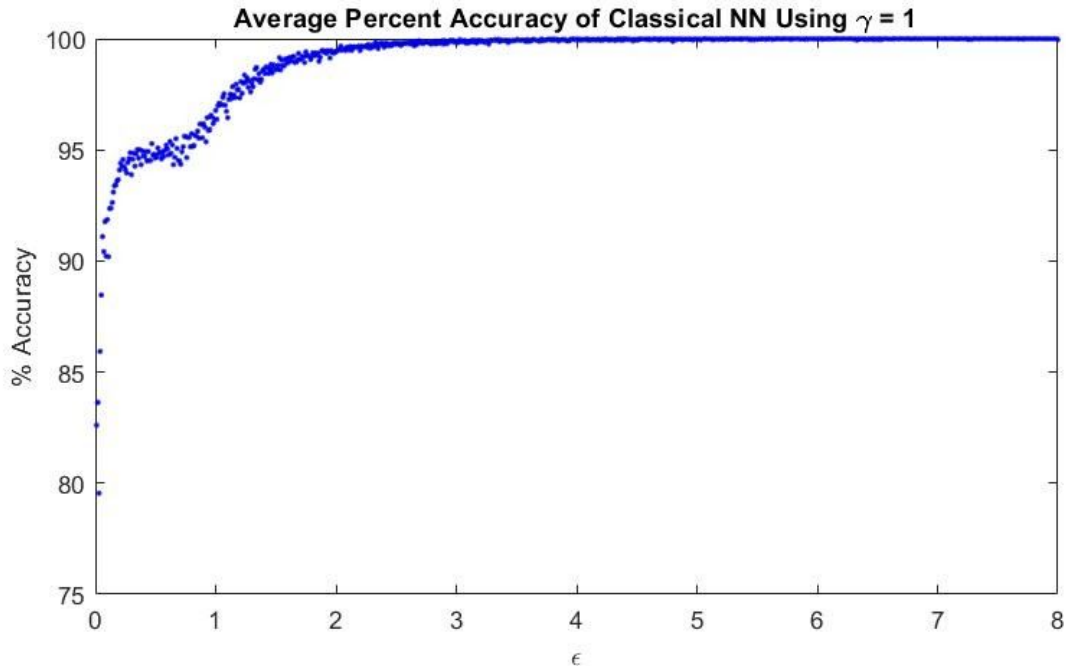

Figure 4.8: Comparing $\varepsilon$ to the average percent accuracy of Classical Artificial Neural Network using the *LS* mechanism where $\gamma = 1$

## 4.5 Deep Learning

Deep Learning is by far one of the most important subsets of machine learning methods. Whereas classical artificial neural networks have two or three hidden layers, deep learning models, or deep neural networks, can contain hundreds of hidden layers. Deep learning models take large labeled training datasets and require high computer processing power. Applications for deep learning include self-driving cars, handwriting recognition, and object classification from images.

We choose to use Deep Learning because of its many implications in modern research, including medical imaging and speech recognition in [10] and [25], and its connections with differential privacy [5]. Deep Learning enables automatic feature extraction during the process of learning, rather than the traditional pre-learning manual feature extraction. The result is a more accurate machine learning model and also a model that can handle vast amounts of data, contrary to methods such as logistic regression, SVM, and SVR. Deep Learning also performs well in classifying large amounts of complex data.

The type of deep neural networks we employed is the Convolutional Neural Network (CNN), which effectively recognizes patterns of dataset in tensor forms. CNN's are a class of Feedforward Artificial Neural Networks and are fully connected networks, meaning the neuron in each layer are connected to each neuron in the next layer.

CNN's extract and learn small patterns in input data primarily through the three operations: convolution, which is performed with a small kernel matrix, ReLu, a non-linear operation that replaces negative values with zero, and pooling, which sizes down the feature maps. In contrast to other types of networks, CNN's input data has greater than or equal to three dimensions. For example, in color image processing, input data has height, width, and depth [22].

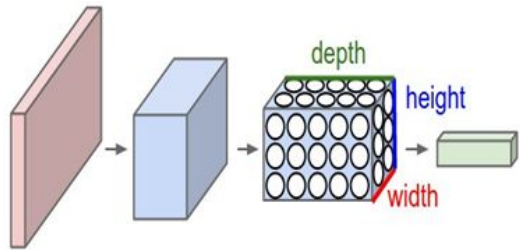

Figure 4.9: Convolutional Neural Network dimensions

![Convolution operations figure]

Figure 4.10: Convolution operations

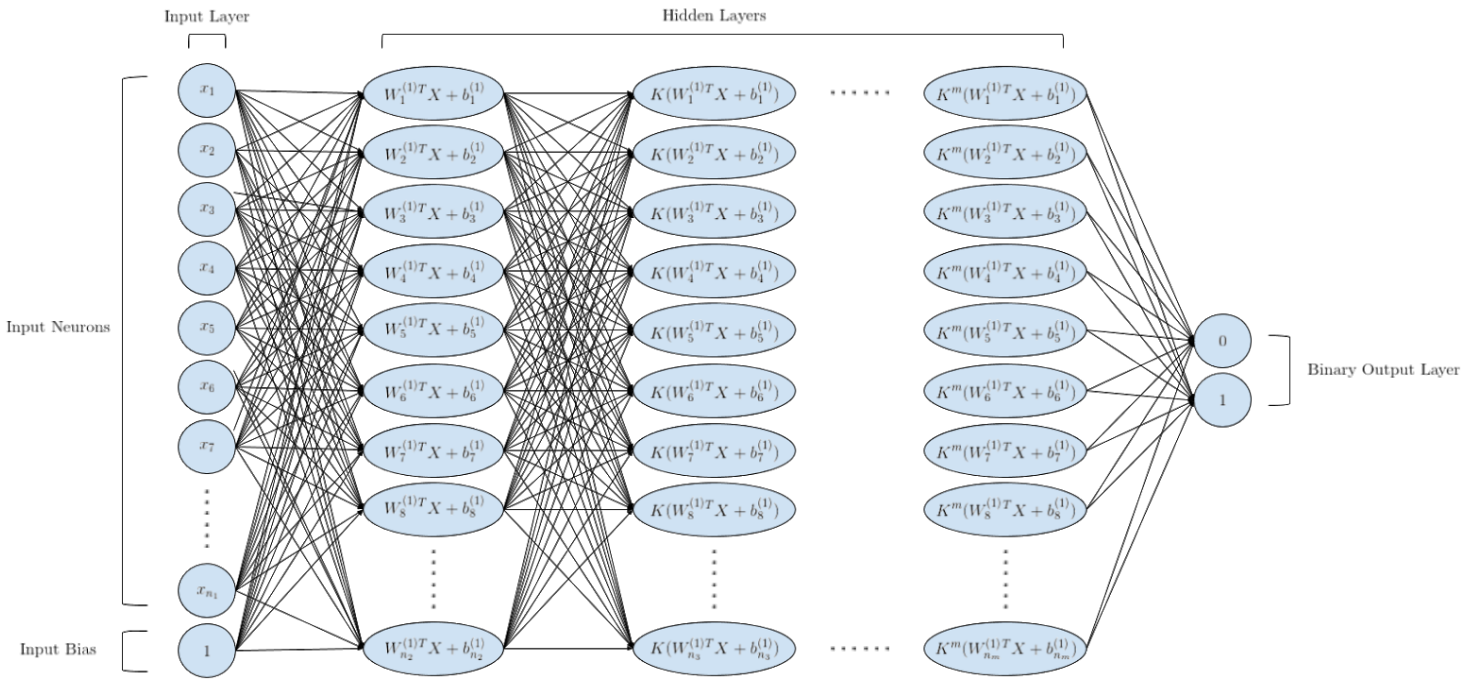

Figure 4.11: A neural network with many hidden layers

For the IPUMS Dataset, we use Google Colab to generate a deep neural network with 5 layers. There are 4 hidden layers with 16, 18, 20, and 24 neuron units in each layer, respectively. We train the network with back-propagation. The activation function for the hidden layers is RELU:

$$R(z) = \max(0, z)$$

and the activation function for the output layer is the sigmoid function:

$$R(z) = \frac{1}{1 + e^{-z}}$$

More details of the Deep Neural Network for the IPUMS Dataset are shown below.

```
Model: "sequential_43"
_________________________________________________________________
Layer (type)                 Output Shape              Param #
=================================================================
dense_211 (Dense)            (None, 16)                144
_________________________________________________________________
dense_212 (Dense)            (None, 18)                306
_________________________________________________________________
dropout_43 (Dropout)         (None, 18)                0
_________________________________________________________________
dense_213 (Dense)            (None, 20)                380
_________________________________________________________________
dense_214 (Dense)            (None, 24)                504
_________________________________________________________________
dense_215 (Dense)            (None, 1)                 25
=================================================================
Total params: 1,359
Trainable params: 1,359
Non-trainable params: 0
_________________________________________________________________
```

Figure 4.12: Deep Neural Network model used for IPUMS Dataset

For the MNIST Dataset, we use the CNN model from Keras [21] to create a convolutional neural network (CNN) with 8 layers. The CNN uses ReLU for the hidden layer activation function, and it uses softmax with 10 classes for the output layer activation function. More details of the CNN are shown below.

```
Model: "sequential_1"
_________________________________________________________________
Layer (type)                 Output Shape              Param #
=================================================================
conv2d_1 (Conv2D)            (None, 25, 25, 32)        320
_________________________________________________________________
conv2d_2 (Conv2D)            (None, 23, 23, 64)        18496
_________________________________________________________________
max_pooling2d_1 (MaxPooling2 (None, 11, 11, 64)        0
_________________________________________________________________
dropout_1 (Dropout)          (None, 11, 11, 64)        0
_________________________________________________________________
flatten_1 (Flatten)          (None, 7744)              0
_________________________________________________________________
dense_1 (Dense)              (None, 128)               991360
_________________________________________________________________
dropout_2 (Dropout)          (None, 128)               0
_________________________________________________________________
dense_2 (Dense)              (None, 10)                1290
=================================================================
Total params: 1,011,466
Trainable params: 1,011,466
Non-trainable params: 0
_________________________________________________________________
```

Figure 4.13: CNN model used for MNIST Dataset

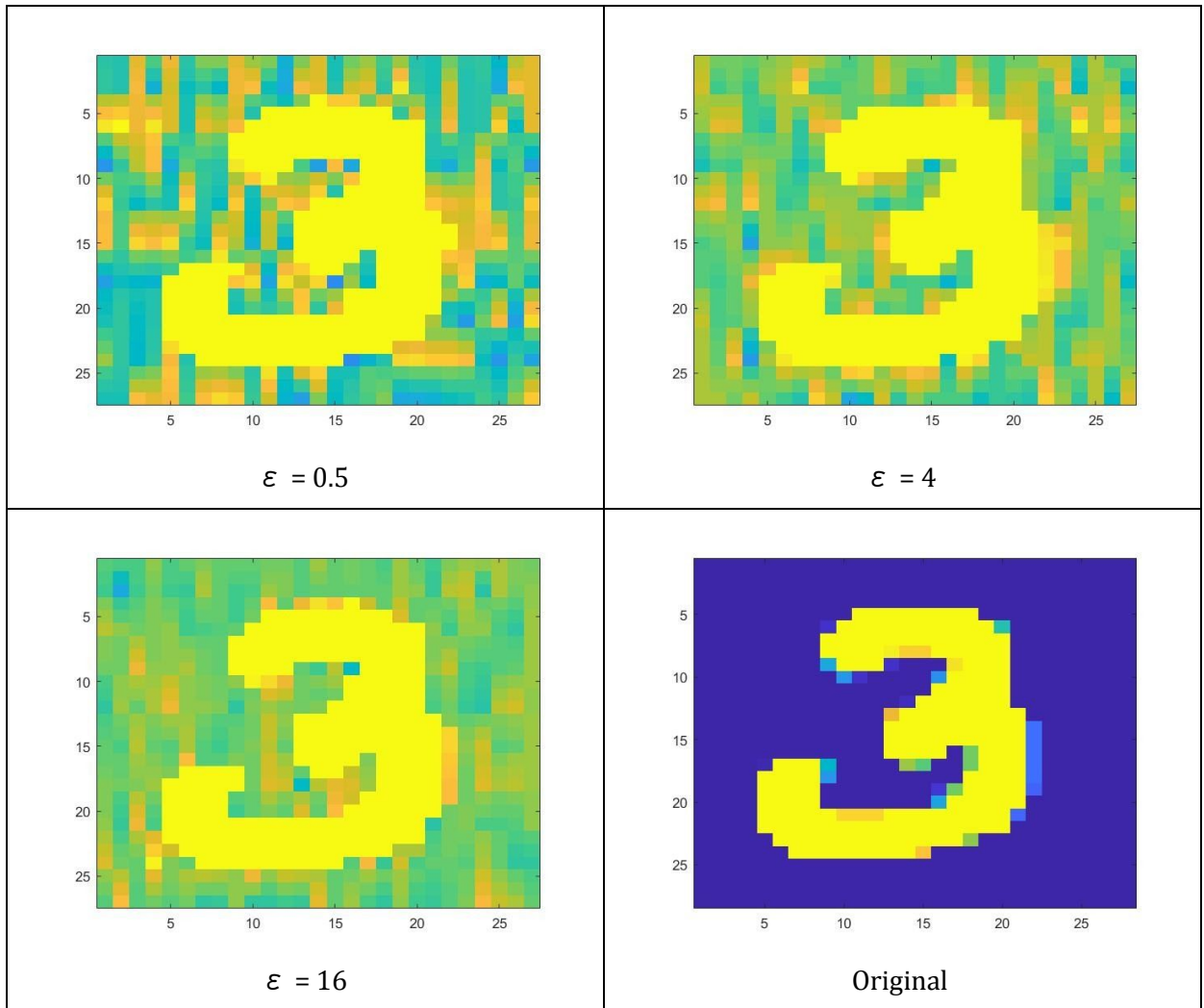

Figure 4.14: MNIST images with different $\varepsilon$ values

As shown in Figure 4.14, the MNIST image becomes less clear at lower $\varepsilon$ values. As more noise is added, the background of the image becomes less smooth.

| LS - Deep NN | | | | | | | | | | | | | | | |
|---|---|---|---|---|---|---|---|---|---|---|---|---|---|---|---|
| $\gamma$ | 0.5 | 1 | 2 | 0.5 | 1 | 2 | 0.5 | 1 | 2 | 0.5 | 1 | 2 | 0.5 | 1 | 2 |
| $\varepsilon$ | 0.5 | | | 1 | | | 2 | | | 4 | | | 8 | | |
| % Accuracy | 94.92 | 94.65 | 94.38 | 94.38 | 94.38 | 94.24 | 94.65 | 94.51 | 95.20 | 94.65 | 94.51 | 94.79 | 95.61 | 95.06 | 95.47 |

| LS+ - Deep NN | | | | | | | | | | | | | | | |
|---|---|---|---|---|---|---|---|---|---|---|---|---|---|---|---|
| $\gamma$ | 0.5 | 1 | 2 | 0.5 | 1 | 2 | 0.5 | 1 | 2 | 0.5 | 1 | 2 | 0.5 | 1 | 2 |
| $\varepsilon$ | 0.5 | | | 1 | | | 2 | | | 4 | | | 8 | | |
| % Accuracy | 94.92 | 95.13 | 95.72 | 95.66 | 97.08 | 98.51 | 98.33 | 99.27 | 99.67 | 99.78 | 99.93 | 99.98 | 99.99 | 99.99 | 100.00 |

| PQ - Convolutional NN | | | | | |
|---|---|---|---|---|---|
| $\varepsilon$ | 0.5 | 1 | 2 | 4 | 8 |
| % Accuracy | 99.09 | 99.08 | 99.11 | 99.11 | 99.14 |

In [5], Abadi *et al.* apply a differentially private stochastic gradient descent (SGD) algorithm to MNIST, using a simple feed-forward neural network with ReLU units and a softmax output layer with 10 classes. Note that [5] guarantees ($\varepsilon$, $\delta$)-differential privacy, which includes an added privacy threshold $\delta$. ($\varepsilon$, $\delta$)-differential privacy is used for mechanisms that do not utilize Laplace noise, and $\varepsilon$-differential privacy is the case where $\delta$ = 0. Thus, we compare our results to [5]'s results from the smallest value of $\delta$ tested, which is $\delta = 10^{-5}$. Several comparisons for different $\varepsilon$ values are shown below:

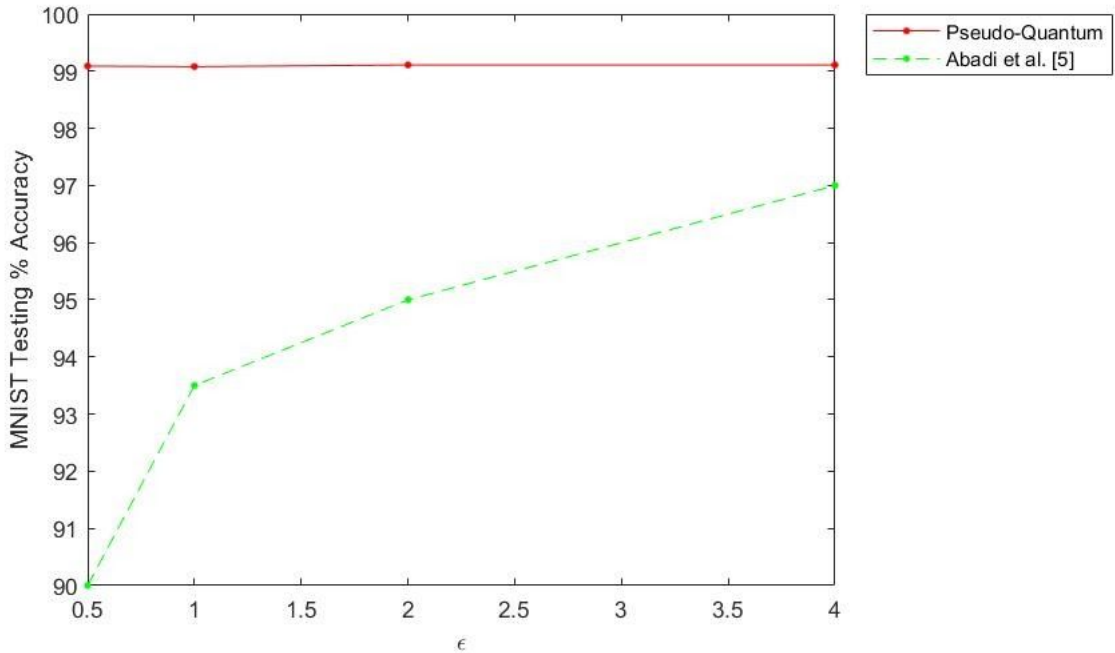

Figure 4.15: Percent accuracies for different $\varepsilon$ values in the pseudo-quantum mechanism compared to Abadi *et al.*'s differentially private SGD in [5]

## 4.6 Differentially Private Imaging

One interesting aspect of the pseudo-quantum mechanism is its ability to introduce complex values to a noisy dataset. In addition to testing our differentially private mechanisms in machine learning environments, we are able to use our pseudo-quantum mechanism to embed complex noise into images.

Recall that the pseudo-quantum noise embedding intensity $\delta$ does not affect $\varepsilon$, the privacy of the dataset. However, $\delta$ affects the statistical accuracy of the data. Higher values of $\delta$ allow more embedding of noise into the dataset, while lower values of $\delta$ diminish the noise's effect.

From Remark 1, the least upper bound of $\delta$ is $\frac{2}{\sqrt{3}} - 1$ for all real mechanism outputs. For the machine learning environments, we set $\delta = 0.1$ and vary $\varepsilon$. However, for embedding noise into images, we have more freedom and can introduce complex values to the image to obscure it further. We do this by surpassing the least upper bound of $\frac{2}{\sqrt{3}} - 1$ for $\delta$. For example, a QR code becomes almost impossible to decipher once complex values are introduced:

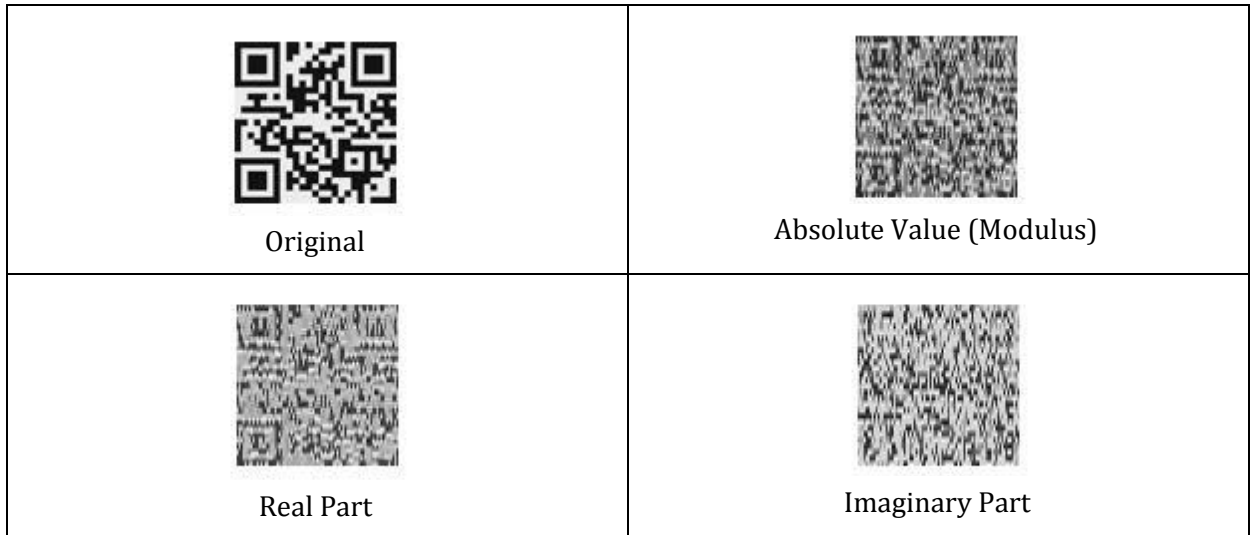

Figure 4.16: Parts of a complex image after noise embedding by the pseudo-quantum mechanism with $\varepsilon = 1$ and $\delta = 0.7$

Notice that the absolute value (modulus) of the image is the best approximation to the original, yet it is not close to being accurate. All three parts of the noisy QR code in Figure 4.16 have become unscannable. We can also test for different values of $\delta$, as shown below.

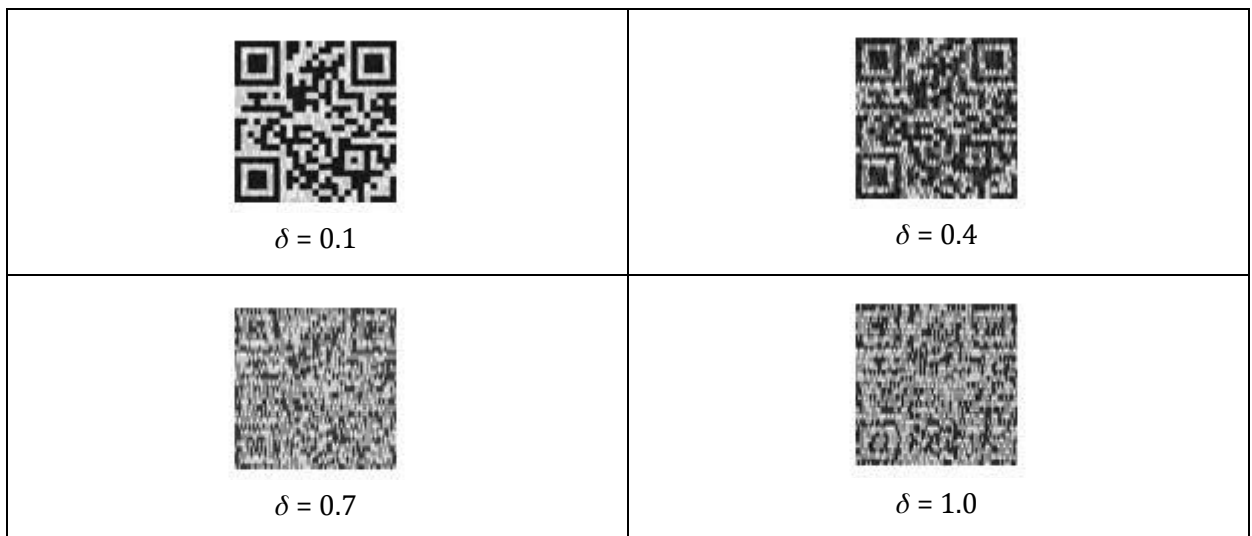

Figure 4.17: Display of the absolute value (modulus) of the image after noise embedding by the pseudo-quantum mechanism with $\varepsilon = 1$

As δ increases, the complex part of the image values increases, and the image becomes more noisy. For δ = 0.7 and 1.0, the QR codes in Figure 4.17 are unscannable.

Although values of δ greater than $\frac{2}{\sqrt{3}} - 1$ allow for more noise addition, they are only applicable to non-machine learning environments, such as imaging, due to the complex parts that are introduced. Because δ is independent from $\varepsilon$, the same level of privacy can still be preserved for all values of δ. Note that all four images in Figure 4.16 have the same privacy, $\varepsilon$ = 1. Low values of δ should be utilized for greater statistical similarity to the original dataset, while high values of δ should be used for large deviation from the original dataset. Future research and testing can be done for larger values of δ.

## 4.7 Results

1. The accuracies for the five machine learning environments are all above 94%, demonstrating that all three mechanisms allow the addition of noise while maintaining the same statistical trends in the original data.

The *LS* and *LS+* mechanisms yield similar accuracies for all machine learning environments except Deep Learning. The relative inaccuracy of the *LS* mechanism in Deep Learning may be due to its smaller-sized training and testing sets compared to those of the *LS+* mechanism. Both the *LS* and *LS+* mechanisms improve in accuracy with increasing $\varepsilon$ values, while the pseudo-quantum mechanism has mostly consistent accuracies.

For *LS* and *LS+*, the accuracies are largely influenced by the $\varepsilon$ parameter. For the vast majority of the trials, $\varepsilon$ = 8 results in the highest accuracies. In Logistic Regression and Classical ANN, both *LS* and *LS+* reach 100% testing classification accuracy when $\varepsilon$ = 8. In addition, as $\gamma$ increases, less noise is added (by Theorem 1). Thus the accuracy increases, as shown as a trend in the *LS* and *LS+* data. Although larger values of $\varepsilon$ provide greater accuracy, they give less privacy to users. Companies should consider the trade-off between higher differential privacy (smaller $\varepsilon$) and higher statistical integrity (larger $\varepsilon$) when choosing the parameter values.

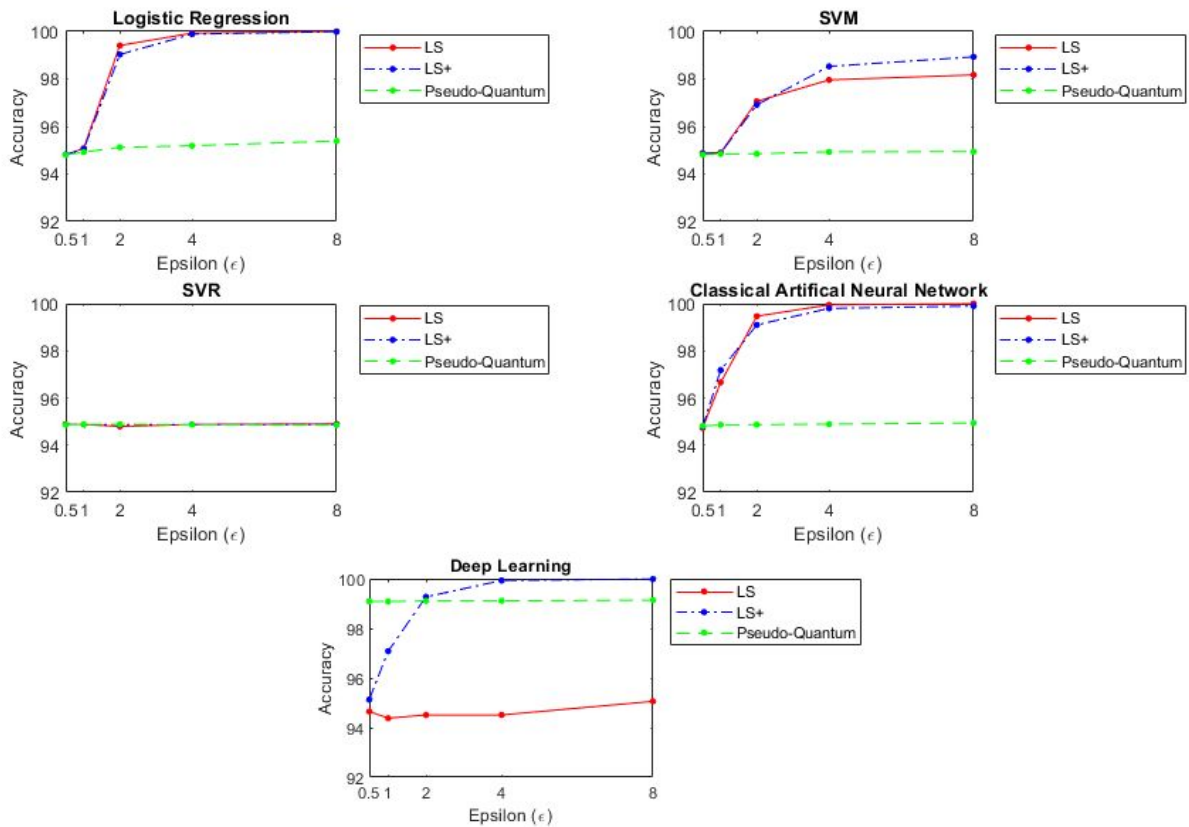

Figure 4.18: Average percent accuracies of mechanisms separated by machine learning environments. $\gamma = 1$ for *LS* and *LS+*, and $\delta = 0.1$ and $\eta = 0$ for the pseudo-quantum mechanism.

2. The pseudo-quantum mechanism consistently scores high accuracies across all five $\varepsilon$ values for all five machine learning environments, with the lowest average accuracy being 94.80% for $\varepsilon = 0.5$ in logistic regression, and the highest being 99.14% for $\varepsilon = 8$ in the Convolutional Neural Network environment. Furthermore, in comparison to Abadi *et. al.*'s differentially private SGD algorithm in [5], the pseudo-quantum mechanism performs exceptionally well (see Figure 4.14). Note that the accuracy achieved in [21] using the same Convolutional Neural Network as ours without any noise addition (no privacy) is 99.25%.

Although the pseudo-quantum mechanism does not give high accuracies relative to the *LS* and *LS+* mechanisms in Logistic Regression, SVM, SVR, and Classical ANN, it is superior in its usage of higher computational power and provides greater data security in comparison to the *LS* and *LS+* mechanisms. Instead of adding noise like the *LS* and *LS+* mechanisms, the pseudo-quantum mechanism embeds noise into the dataset in the wavelet domain. As mentioned in Section 3.4.3, the probability of decoding the noise added in the pseudo-quantum mechanism is $2^{-mn/3}$ for $\eta = 0$, which is highly improbable.

Varying $\delta$ in the pseudo-quantum mechanism does not affect the privacy of the dataset, but it controls the numerical deviation of the noisy dataset from the original dataset, as shown in Section 4.6. High values of $\delta$ should only be used in non-machine learning environments, such as imaging, because they allow for complex values in the noisy dataset. Data privacy managers should pay close attention to the increasing obscurity of the noisy dataset as they increase the $\delta$ values.

3. When choosing between the three mechanisms, data scientists should consider using *LS* or *LS+* for binary classification models (such as Linear Regression and Support Vector Machine) and shallow neural networks, while they should use the pseudo-quantum for accuracy in larger datasets and more complex machine learning environments (such as Convolutional Neural Networks).

# 5 Conclusions and Future Research

Instead of spending time and resources attempting to develop encryption methods that do not necessarily guarantee $\varepsilon$-differential privacy or protection from adversary attacks, companies can use privacy-preserving mechanisms to protect personal data. Our three proposed mechanisms encrypt noise into data after DMWT, which presents many advantages over other methods. First, input perturbation allows easy integration of our mechanisms into existing data environments. In addition, removing the noise becomes practically impossible as long as the adversary does not know the exact noise function, which is unique to each scenario, as the function depends on the set of inputs being changed. If put into use, companies will find our input perturbation mechanisms slightly more complicated but more effective.

Our mechanisms add sufficient noise to achieve $\varepsilon$-differential privacy while still preserving overall statistical trends within the dataset. Setting $\gamma = 1$, $\delta = 0.1$, and $\eta = 0$, we achieve the following average accuracies for our three mechanisms for $\varepsilon = 1$ tested in the five machine learning environments:

|  | Logistic Regression | SVM | SVR | Classical ANN | Deep Learning |
| --- | --- | --- | --- | --- | --- |
| *LS* | 95.08% | 94.90% | 94.88% | 96.66% | 94.38% |
| *LS+* | 95.05% | 94.88% | 94.87% | 97.18% | 97.08% |
| Pseudo-Quantum | 94.92% | 94.84% | 94.88% | 94.86% | 99.08% |

For all five machine learning methods, the models correctly predict the label variables with average accuracies greater than 94%. The results show that companies can employ wavelet transformations and still be able to analyze the dataset for correlations and trends. In addition, all three mechanisms use input perturbation, which is the addition of noise directly into the dataset. This allows for easy integration with existing data environments, such as MySQL and Vertica, because the mechanisms do not need to be modified for each data management system.

The *LS+* mechanism is applicable to dynamic datasets, which are datasets that are constantly increasing in size. As a result, companies could store two separate datasets—private and public—where the public dataset is updated with every new batch of samples added to the private dataset and still preserves differential privacy through use of *LS+*. Furthermore, the partitioning aspect of the *LS+* mechanism allows for parallel computing, cutting down on computational time even more. In the future, more complex deep learning methods should be used for *LS* and *LS+* in the future to increase our mechanisms' accuracy.

The pseudo-quantum mechanism stands as the most advanced method out of our three mechanisms because of its higher computational power. Additionally, it provides the most consistent accuracies for all $\varepsilon$ values. We see the future of data privacy as a collaborative effort between engineers and data scientists working to preserve differential privacy for the safety of the greater community. The pseudo-quantum mechanism is one of the first steps towards that goal.

In the future, more research and testing can go into improving the pseudo-quantum mechanism to withstand greater noise and enabling its use for dynamic datasets and parallel computing like the *LS+* mechanism. As our mechanisms require long processing times for large datasets, another priority would be to decrease the computation time by finding new ways to create and store large wavelets and perform the processes. Additionally, a major goal can be to incorporate randomized response, which is outlined in [14], in the three mechanisms so that the original dataset does not have to be stored to provide high statistical integrity.

# 7 Acknowledgment

We would like to acknowledge the contributions of each member of the team. Kenneth was the first author who led and kept the team on task, created and coded the three mechanisms, coded most of the machine learning environments in MATLAB and Python, and wrote the sections related to the mechanisms and their proofs. Tony worked on writing about support vector regression, delving into its mathematical steps. Kenneth ran a significant amount of trials and collected their results, contributing substantial time and effort into the research.

We would also like to extend our special thanks to our advisor, Dr. Wang, who has provided us with class time to learn about wavelets and machine learning, innumerable hours of assistance, and many resources. Only with Dr. Wang's aid and the Western Connecticut State University summer courses and their computer laboratory facilities were we able to complete the research. We are also very grateful for our assistant, Meera Sharma, who helped us connect our research to the wavelet transform, and for our mentors Ralph Venezia and Hieu Nguyen, who generously introduced us to the basics of MATLAB and Python. We have come to learn many aspects of mathematics, computer science, and teamwork in a limited time.

Link to all data and mechanism functions:
http://www.mediafire.com/file/mo5h59saqe0r21k/Differential_Privacy_2019_-_Kenneth_Choi_and_Tony_Lee.zip/file